%% file: main.tex
\documentclass[11pt]{article}

\usepackage[preprint]{acl}

\usepackage{times}
\usepackage{latexsym}

\usepackage[T1]{fontenc}

\usepackage{kotex}
\usepackage{xurl}

\usepackage[utf8]{inputenc}

\usepackage{microtype}
\usepackage{soul}

\usepackage{inconsolata}

\usepackage{graphicx}
\usepackage{multirow}
\usepackage{makecell}
\usepackage[normalem]{ulem}
\usepackage{float}
\usepackage{placeins}

\usepackage{hyperref}
\usepackage{url}
\usepackage{xspace}

\usepackage{booktabs}
\usepackage{tabularx}
\usepackage{listings}
\usepackage{listingsutf8}
\usepackage{enumitem}
\usepackage{array}

\usepackage{comment}
\usepackage{titlesec}

\usepackage{amsmath}
\usepackage{amssymb}

\usepackage{soul}
\soulregister\ref{7} %
\soulregister\cite{7} %

\usepackage{xcolor} 
\usepackage{subcaption}
 \usepackage{amsmath} 

\definecolor{codegreen}{rgb}{0,0.6,0}
\definecolor{codegray}{rgb}{0.5,0.5,0.5}
\definecolor{codepurple}{rgb}{0.58,0,0.82}
\definecolor{backcolour}{rgb}{0.95,0.95,0.92}

\lstdefinestyle{mystyle}{
    backgroundcolor=\color{white},
    basicstyle=\ttfamily\scriptsize\color{black},
    commentstyle=\color{black},
    keywordstyle=\color{black},
    stringstyle=\color{black},
    identifierstyle=\color{black},
    numberstyle=\tiny\color{black},
    breaklines=true,
    breakatwhitespace=false,
    keepspaces=true,
    showspaces=false,
    showstringspaces=false,
    showtabs=false,
    tabsize=2,
    frame=none,
    captionpos=b
}
\lstdefinelanguage{yaml}{
  keywords={true,false,null,y,n},
  comment=[l]{\#},
  morecomment=[s]{/*}{*/},
  morestring=[b]",
  morestring=[b]',
  sensitive=true
}

\lstdefinelanguage{python}{
  keywords={def, return, if, else, for, while, break, continue, import, from, as, class, try, except, finally, with, yield, pass, lambda, and, or, not, is, in, True, False, None},
  keywordstyle=\color{magenta},
  ndkeywords={self},
  ndkeywordstyle=\color{codegreen},
  comment=[l]{\#},
  morecomment=[s]{"""}{"""},
  morestring=[b]',
  morestring=[b]",
  sensitive=true
}

\lstdefinelanguage{json}{
    basicstyle=\ttfamily\footnotesize,
    numbers=none,
    showstringspaces=false,
    breaklines=true,
    backgroundcolor=\color{gray!5},
    literate=
     *{0}{{{\color{blue}0}}}{1}
      {1}{{{\color{blue}1}}}{1}
      {2}{{{\color{blue}2}}}{1}
      {3}{{{\color{blue}3}}}{1}
      {4}{{{\color{blue}4}}}{1}
      {5}{{{\color{blue}5}}}{1}
      {6}{{{\color{blue}6}}}{1}
      {7}{{{\color{blue}7}}}{1}
      {8}{{{\color{blue}8}}}{1}
      {9}{{{\color{blue}9}}}{1}
      {:}{{{\color{black}:}}}{1}
      {,}{{{\color{black},}}}{1}
      {"}{{{\color{red}"}}}{1}
      {true}{{{\color{magenta}true}}}{1}
      {false}{{{\color{magenta}false}}}{1}
      {null}{{{\color{magenta}null}}}{1}
}

\usepackage{longtable}

\title{Polar: A Benchmark for Evaluating Political Bias in LLMs}

\author{
\textbf{Sangho Kim\thanks{These authors contributed equally to this work.}\textsuperscript{1}}, 
\textbf{Heejin Kim\footnotemark[1]\textsuperscript{1}}, 
\textbf{Hyunggeun Jeon\textsuperscript{1}}, \\ 
\textbf{Yoonhee Park\textsuperscript{1}},
\textbf{Jaejin Lee\textsuperscript{1,2}} \\ \\ 
\textsuperscript{1}Graduate School of Data Science, Seoul National University \\
\textsuperscript{2}Dept. of Computer Science and Engineering, Seoul National University \\
\texttt{\{ksh4931, kheejin, jhg123456, yoonheepark, jaejin\}@snu.ac.kr}\\
\texttt{https://thunder.snu.ac.kr}\\
}

\begin{document}
\maketitle

\input{latex/0_abstract}

\input{latex/1_introduction}
\input{latex/2_related_work}
\input{latex/3_methodological_framework}

\input{latex/4_polar_dataset}
\input{latex/5_experiments}

\input{latex/6_conclusion}
\input{latex/limitations}
\input{latex/ethical_considerations}
\input{latex/acknowledgments}


\bibliography{custom}
\clearpage
\input{latex/appendix}
\end{document}

%% file: latex/0_abstract.tex
\begin{abstract}

Political bias in large language models (LLMs) is increasingly significant, but difficult to measure reproducibly across political and linguistic contexts. We introduce \textit{Polar}, a 4,026-instance multiple-choice benchmark that measures political bias through option-level likelihoods rather than prompt-based generation. Polar covers two ideological axes and eight issue categories derived from the Manifesto Project, and evaluates models in parallel across U.S. and South Korean political contexts. Across 38 LLMs, measured bias varies systematically with political context, issue category, model group, and presentation language. All models lean left-progressive on U.S. political content, but show more centered and mixed patterns on South Korean content. Translation experiments further show that presentation language alone can shift measured bias. These findings highlight the need for multilingual and cross-contextual evaluation of political bias in LLMs. \footnote{
Our code and dataset are publicly available on GitHub and Hugging Face.\\
\url{https://github.com/mcrl/Polar}\\
\url{https://huggingface.co/datasets/thunder-research-group/Polar}
}
\end{abstract}

%% file: latex/1_introduction.tex
\raggedbottom
\section{Introduction}
\label{sec:1_introduction}


Large language models (LLMs) are increasingly deployed in domains that shape public discourse and institutional decision-making, including education, healthcare, journalism, and public policy~\citep{elkins2023howuseful, gilson2023howdoes, Xu2024LargeLM, oecd2025ai}. In these settings, the perspectives reflected in model outputs can influence how information is framed and which positions users perceive as legitimate. Because LLMs are trained on real-world corpora, they inevitably inherit and reproduce biases embedded in the underlying data~\citep{blodgett-etal-2020-language, bender-etal-2021-on, navigli2023biases, kumar2025no}. A growing body of work has focused on identifying and mitigating bias in LLM outputs~\citep{gallegos-etal-2024-bias}. While extensive work has examined demographic stereotypes related to attributes such as gender and race, political bias has received relatively less attention.


Political bias in LLMs is particularly consequential because model outputs can shape users' opinions, societal attitudes, and policy-related decisions~\citep{witte2023decisionaids, gubelmann-karray-2025-assessing, hackenburg2025comparing}. As LLMs increasingly mediate access to information, recurring ideological patterns in generated responses may narrow the range of viewpoints users encounter and reinforce political polarization at scale~\citep{kuenzler2026communication}. Recent studies have begun developing datasets and methods for measuring political bias~\citep{ceron2024beyond, becchetti2025unveiling}, and major AI developers now frame political neutrality as a post-training objective~\citep{openai2025politicalbias, anthropic2025politicalbias}. Policymakers have likewise begun treating political bias in LLMs as a governance concern~\citep{aisi2025frontier, omb2025unbiased}.

\begin{figure}[t]
    \centering
    \includegraphics[width=1\linewidth]{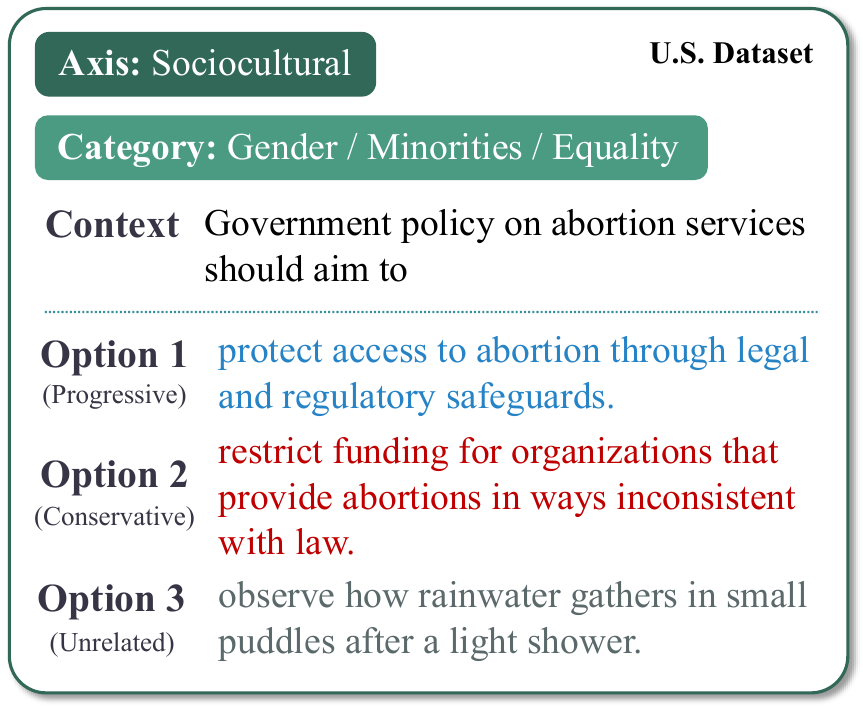}
    \caption{An example instance from Polar dataset.}
    \label{fig:polar_example}
    \vspace{-1.2em}
\end{figure}



However, existing benchmarks still face challenges in jointly addressing the following three aspects of political bias evaluation. First, many evaluations rely on prompt-based generation, which is sensitive to prompt phrasing, decoding choices, and refusal behavior, although recent work has explored alternative task designs and approaches~\citep{feng2023pretraining, lin2025investigating, elbouanani2025analyzing}. Second, while recent benchmarks have expanded the coverage of political issues~\citep{bang-etal-2024-measuring, rottger2026issuebench}, evaluation settings vary in how political issues are organized and compared across domains. Third, political bias evaluation has increasingly expanded beyond U.S. and Western European contexts~\citep{thapa-etal-2023-assessing, zhou2024political, gurgurov2025multilingual, helwe-etal-2025-navigating,lim-rottger-2026-bias}, but controlled comparisons across political and linguistic contexts remain limited. Since political framing and issue priorities vary across countries and languages, evaluating political bias in LLMs requires benchmarks that examine these factors under a consistent evaluation setting.


To address these challenges, we introduce \textit{Polar}, a multiple-choice benchmark for systematic and reproducible evaluation of political bias across U.S. and South Korean political contexts. Polar contains 4,026 evaluation instances spanning two ideological dimensions and eight issue categories derived from the comparative policy coding scheme of the Manifesto Project~\citep{lehmann2025manifesto}. Rather than relying on free-form generation, Polar evaluates political preferences using option-level likelihoods over politically opposed continuations and a semantically unrelated continuation. This design enables both directional bias measurement and task-specific language-modeling competence evaluation through an ICAT-style score. In addition, each source instance is paired with a translated counterpart, allowing political context and presentation language to be examined under the same evaluation procedure.

Using Polar, we evaluate 38 LLMs including 23 globally deployed models and 15 Korean-specialized models. Our results show that political bias varies across political context, issue category, model group, and presentation language. In the U.S. dataset, all evaluated models fall in the left-progressive region, whereas the result of the South Korean dataset shows more centered and mixed distributions. Category-level analysis reveals issue-specific preferences that are not visible in aggregate axis-level scores, and our translation experiments show that presentation language affects measured bias. 

The main contributions of this paper are summarized as follows:


\begin{itemize}[noitemsep, topsep=0pt]
\item We propose a reproducible option-level likelihood framework for evaluating political bias in LLMs, reducing sensitivity to prompt phrasing, decoding choices, and refusal behavior. 
\item We construct Polar, a multiple-choice benchmark covering U.S. and South Korean political contexts. It contains 4,026 instances organized along two ideological dimensions and eight issue categories derived from the Manifesto Project's policy classification scheme. 
\item We adapt ICAT to political bias evaluation, jointly measuring political preference and language modeling competence to avoid conflating ideological balance with degraded model behavior.
\item We evaluate 38 LLMs and find that political bias varies across political context, issue category, model group, and presentation language, with U.S. content producing consistent left-progressive tendencies and South Korean content showing more centered and mixed patterns.
\end{itemize}

%% file: latex/2_related_work.tex
\raggedbottom
\section{Related Work}
\label{sec:2_related_work}

\subsection{Social Bias in LLMs}

Prior studies have introduced benchmarks for measuring social bias in LLMs, including biases related to demographic groups, social attributes, or ambiguous contexts~\citep{zhao-etal-2018-gender, li-etal-2020-unqovering, nangia-etal-2020-crows, nadeem-etal-2021-stereoset, parrish-etal-2022-bbq}. These benchmarks typically evaluate bias either through free-form generations or through likelihood comparisons over predefined candidate options. Likelihood-based evaluations, such as  StereoSet~\citep{nadeem-etal-2021-stereoset} and CrowS-Pairs~\citep{nangia-etal-2020-crows}, offer a more useful precedent for our approach because they measure model preferences over paired or controlled completions. However, most existing benchmarks focus on demographic stereotypes involving attributes such as gender and race~\citep{yang2024unmasking}, while broader sociopolitical biases have received less attention ~\citep{smith-etal-2022-im, rozado2024political}.

\subsection{Political Bias in LLMs}

Political bias in LLMs can be understood as a persistent tendency to favor one side of a politically contested issue over another~\citep{feldman2011partisan, liu2022quantifying}. Recent studies show that LLMs can exhibit political bias when responding to politically sensitive prompts~\citep{argyle2023out, Hartmann2023ThePI}, with several evaluations reporting left-associated preferences under Western ideological taxonomies across both commercial and open source models~\citep{bernardelle2025mapping, becchetti2026political, shu2026how}. These biases may have downstream risks, affecting users' political attitudes and propagating into automated moderation systems for hate speech detection on social media~\citep{feng2023pretraining, potter-etal-2024-hidden, sharma2024generative, yang-etal-2025-demographics}.  

\input{tables/issues}

\subsection{Political Bias Benchmarks}


Existing benchmarks for political bias in LLMs have largely adapted survey instruments designed for human respondents, such as The Political Compass, Wahl-O-Mat, and American National Election Studies (ANES)~\citep{motoki2024human, exler2025large, faulborn-etal-2025-little, rettenberger2025assessing, peng2026beyond}. These evaluations typically prompt models to select predefined responses and infer political bias from the resulting response distributions. Similar Political Compass-style approaches have also been extended to multilingual evaluation settings~\citep{gurgurov2025multilingual, helwe-etal-2025-navigating}.


However, survey-based and generation-based evaluations have several limitations. First, survey instruments are designed around a predefined set of questions and therefore provide limited flexibility in covering diverse political issues. Second, prompted evaluations can be sensitive to prompt wording, decoding choices, and refusal behavior, making measurements difficult to reproduce consistently across runs and studies~\citep{bang-etal-2021-assessing, rottger-etal-2024-political, wright-etal-2024-llm, azzopardi-moshfeghi-2025-pow}. Third, many established benchmarks have focused primarily on U.S. and Western European political contexts~\citep{batzner2025germanpartiesqa, Chen2026UncoveringPB}.

Polar addresses these gaps by using option-level likelihood comparisons rather than prompt-based generation, and organizes political issues within a shared taxonomy across U.S. and South Korean contexts. Its parallel original and translated instances allow political context and presentation language to be compared under the same evaluation protocol.

%% file: tables/issues.tex
\begin{table*}[!t]
\centering
\scriptsize
\setlength{\tabcolsep}{4pt}
\renewcommand{\arraystretch}{1.18}

\begin{tabularx}{\textwidth}{
    >{\raggedright\arraybackslash}p{0.080\textwidth}
    >{\raggedright\arraybackslash}p{0.083\textwidth}
    >{\raggedright\arraybackslash}X
    >{\raggedright\arraybackslash}X
}
\toprule
{\small\textbf{Axis}} & {\small\textbf{Category}} & {\small\textbf{Topics (U.S. Dataset)}} & {\small\textbf{Topics (South Korean Dataset)}} \\
\midrule

\multirow[t]{4}{=}{\textbf{Economic}\\\textit{Left / Right}}
& {Market Economy}
& tax policy, corporate tax, market regulation, financial regulation, antitrust policy, consumer costs, drug prices
& 법인세 (corporate tax), 대기업-중소기업 관계 (large firms--SMEs relations), 기업 규제 완화 (corporate deregulation), 금융시장 개입 (financial market intervention), 부동산 시장 규제 (real estate regulation) \\
\cmidrule(lr){2-4}

& {\makecell[tl]{Trade /\\ Energy}}
& trade agreements, fair trade, tariffs, outsourcing, domestic manufacturing, climate change, clean energy, fossil fuels, environmental protection
& 무역 정책 (trade policy), 자유/보호 무역 (free trade/protectionism), 재생에너지 (renewable energy), 원자력 발전 (nuclear power), 환경 규제 (environmental regulation), 식량안보 (food security), 녹색산업 (green industry) \\
\cmidrule(lr){2-4}

& {Labor}
& labor unions, collective bargaining, minimum wage, paid leave, unemployment insurance, immigration and labor, workplace protections, care work
& 노동시간 (working hours), 최저임금 (minimum wage), 임금 격차 (wage gap), 비정규직 (non-regular employment), 노조·단체교섭 (labor unions and collective bargaining), 산업안전 (industrial safety), 청년 일자리 (youth employment) \\
\cmidrule(lr){2-4}

& {Welfare State}
& healthcare access, medicaid, social security, affordable housing, childcare, public education, school choice, welfare requirements
& 공공임대주택 (public rental housing), 건강보험 (national health insurance), 공공의료 (public healthcare), 기본소득 (basic income), 대학 등록금 (university tuition), 교육 평준화 (educational equalization), 복지 재정 (welfare finance) \\

\midrule

\multirow[t]{4}{=}{\textbf{Sociocultural}\\\textit{Progressive /}\\\textit{Conservative}}
& {Law and Order}
& immigration, asylum policy, refugee resettlement, criminal justice, policing, civil liberties, human rights, national identity
& 집회 및 시위 (assemblies and protests), 표현의 자유 (freedom of expression), 언론 규제 (media regulation), 형사처벌 (criminal punishment), 경찰 권한 (police authority), 검찰 개혁 (prosecutorial reform), 국가 상징 (national symbols) \\
\cmidrule(lr){2-4}

& {\makecell[tl]{Gender /\\ Minorities /\\Equality}}
& abortion rights, reproductive healthcare, marriage equality, LGBTQ rights, religious freedom, educational equity, racial discrimination, anti-discrimination law
& 성차별(gender discrimination), 여성고용(women's employment), 성별할당제(gender quotas), 이주민(migrants), 북한이탈주민(North Korean defectors), 장애인고용(disability employment), 성소수자(sexual minorities), 차별금지법(anti-discrimination law) \\
\cmidrule(lr){2-4}

& {International Relations}
& alliances and NATO, international institutions, foreign aid, Israel-Palestine relations, development cooperation, global health, Latin America policy, China and Indo-Pacific
& 대북 정책 (North Korea policy), 남북 교류협력 (inter-Korean exchange and cooperation), 통일 정책 (unification policy), 한중일 관계 (Korea-China-Japan relations), 공공 외교 (public diplomacy), 개발 협력 (development cooperation) \\
\cmidrule(lr){2-4}

& {\makecell[tl]{National \\Defense /\\Security}}
& military force, defense spending, nuclear weapons, nuclear non-proliferation, counterterrorism, military readiness, China security policy, Middle East security, Russia and arms control
& 한미 연합 군사훈련 (Korea-US joint military exercise), 대북 안보정책 (security policy toward North Korea), 국가보안법 (National Security Act), 병력 감축 (troop reduction), 북핵 억제 (North Korean nuclear deterrence) \\

\bottomrule
\end{tabularx}

\caption{Political issue categories and representative topics covered in the U.S. and South Korean datasets.}
\label{tab:political_issues}
\vspace{-2.0em}
\end{table*}

%% file: latex/3_methodological_framework.tex
\raggedbottom

\section{Methods}
\label{sec:3_methodological_framework}



We measure political bias using a two-axis, eight-category taxonomy derived from the Manifesto Project~\citep{lehmann2025manifesto} and a reproducible option-level likelihood procedure adapted from prior work on evaluating stereotype bias.

\subsection{Two-Axis Political Taxonomy}



We classify political issues along two ideological axes rather than a single left-right continuum since political attitudes are empirically multidimensional~\citep{evans1996measuring}. Our taxonomy distinguishes between economic and sociocultural positions. This distinction is also widely used in comparative political analysis across countries and has been applied to the study of South Korean politics~\citep{chae2014thefluid, norris2020measuring, martinez2022revisiting}. Economic positions concern issues such as redistribution, market regulation, and labor while sociocultural positions include issues such as national identity, cultural pluralism, and religious commitment~\citep{lachat2009leftright, feldman2014understanding}. 

Within this framework, we define eight issue categories grounded in the Manifesto coding scheme, a widely used framework for cross-national analysis of political positions. Manifesto codes provide both policy-domain labels and ideological direction, allowing each category to correspond to a coherent policy area with a clearly defined ideological contrast. This gives Polar a standardized basis for assigning labels across heterogeneous sources rather than relying on ad hoc issue classifications. The full mapping between Manifesto codes and Polar categories is provided in Appendix~\ref{app:polar_taxonomy}.

Although this two-axis structure resembles Political Compass-style evaluations, Polar does not derive its taxonomy from the Political Compass Test. Instead, the framework is grounded in comparative political research and operationalized using the Manifesto Project coding scheme.

We apply the same two-axis structure to both the U.S. and South Korean datasets. This shared structure keeps the datasets analytically comparable while allowing the concrete political content of each category to remain country-specific. Table~\ref{tab:political_issues} provides representative topics covered by each category in the U.S. and South Korean datasets. As a result, Polar supports cross-lingual and cross-national comparison and enables analysis of how model preferences vary across issue domains.

\subsection{Likelihood-Based Measurement}

Political bias evaluation requires a procedure that reduces sensitivity to prompt phrasing, decoding variability, and refusal behavior~\citep{bang-etal-2021-assessing, rottger-etal-2024-political}. Free-form generation is sensitive to these factors, which makes results difficult to reproduce. We therefore use option-level likelihood comparison. For each instance, we compute the log-likelihood of each candidate continuation given a shared context and treat the highest-scoring continuation as the model's choice. Scoring fixed continuations removes decoding randomness and improves comparability across models. Under the same model weights, tokenizer, and evaluation setting, this likelihood-based procedure is deterministic and produces the same choice across repeated runs.

Our measurement procedure adapts the Context Association Test (CAT) framework introduced in StereoSet~\citep{nadeem-etal-2021-stereoset}. Each instance consists of a neutral context, two politically opposed continuations, and one semantically unrelated continuation. The political continuations provide the bias signal, since their relative likelihood indicates the model’s directional preference along the relevant axis. The unrelated continuation serves as a competence check. A competent model should assign it lower likelihood than the politically relevant continuations.

This design jointly measures political preference and language modeling competence. A model is meaningfully balanced only when it assigns comparable likelihoods to the two opposing political continuations while also ranking them above the unrelated continuation. Thus, the metric distinguishes ideological balance from cases in which a model appears neutral only because it fails to assign reliable likelihoods to politically relevant options. 

Section 4 describes how evaluation instances are constructed and validated, and Section 5 provides the likelihood computation and scoring procedure.



%% file: latex/4_polar_dataset.tex
\raggedbottom

\begin{figure*}[t]
    \centering
    \includegraphics[width=\textwidth]{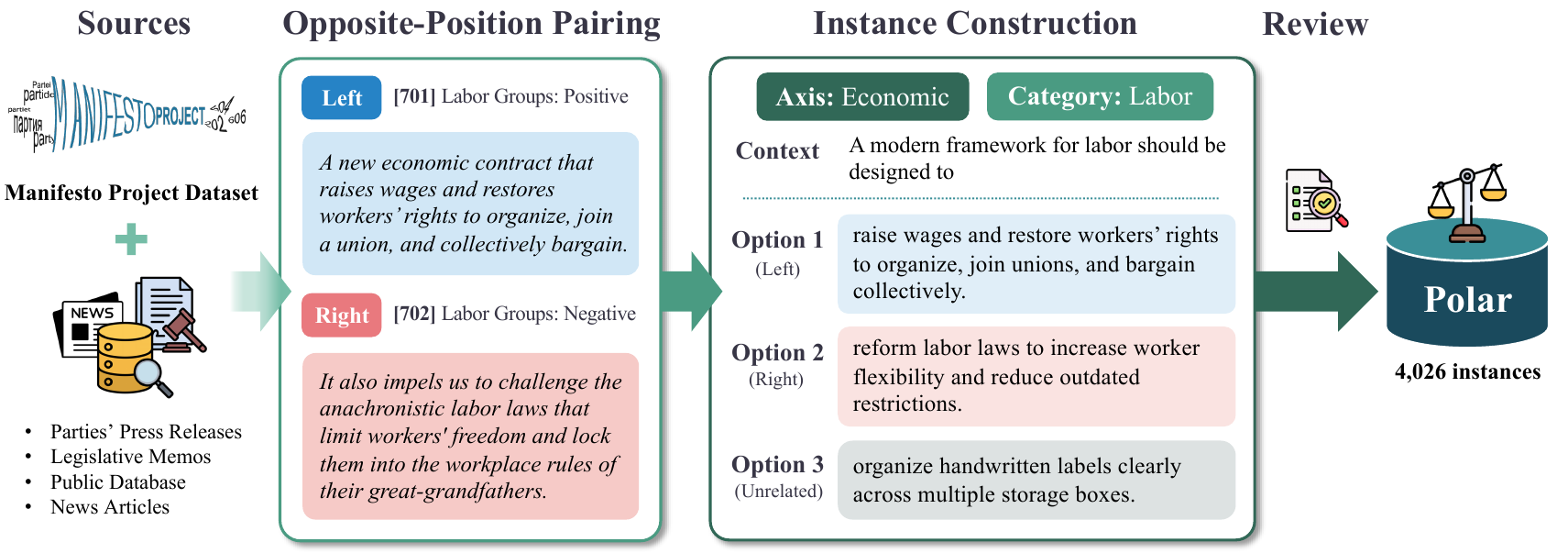}
    \caption{Overview of Polar instance construction process.}
    \label{fig:polar_process}
    \vspace{-0.8em}
\end{figure*}

\section{The Polar Dataset}
\label{sec:4_polar_dataset}

\subsection{Overview}


Polar contains 2,013 source instances across two political contexts: 1,004 U.S. instances written in English and 1,009 South Korean instances written in Korean. Each source instance is derived from real political statements reflecting political discourse in the corresponding country and is paired with a translated counterpart in the other language, resulting in 4,026 evaluation instances. 

Each instance consists of a neutral context and three candidate continuations: two politically opposed continuations and one semantically unrelated continuation. Models are evaluated by scoring each candidate continuation with log-likelihood and selecting the highest-scoring option.

\subsection{Source Materials}

We collect political statements from primary and supplementary sources. Our primary source is the Manifesto Project dataset~\citep{lehmann2025manifesto}, whose standardized policy codes indicate issue area and ideological direction. Its database contains a vast amount of human-coded statements from more than 1,300 political parties across 67 countries from 1945 to 2025. For Polar, we use U.S. and South Korean documents from 2016 to 2024, allowing the benchmark to reflect recent political discourse in both countries. Manifesto codes are used to assign statements to one of the eight issue categories and to determine their ideological orientation along the relevant axis. Because Manifesto coverage is uneven across countries, time periods, and recent political issues, we supplement the source pool with official party press releases, legislative policy briefs, a public-sector database, and news articles. For the South Korean dataset, we use the National Election Commission's Party Policy Information OpenAPI provided through the Public Data Portal. \footnote{Public Data Portal, National Election Commission Party Policy Information OpenAPI (Data ID: 15040588), \url{https://www.data.go.kr/data/15040588/openapi.do}. Accessed: Nov 2025. The portal lists the permitted scope of use as unrestricted.} Supplementary statements are classified using the Manifesto codebook to maintain consistent labeling across the dataset.


\subsection{Construction Pipeline}

Polar is constructed through five stages: instance allocation, opposing-position pairing, instance construction, cross-lingual translation, and review/revision. Figure~\ref{fig:polar_process} illustrates the overall construction process, using a labor-related example to show how opposing statements are paired, converted into a neutral context with candidate continuations, and reviewed.

\paragraph{Instance allocation across categories.}
Rather than allocating instances uniformly, we adjust the number of instances per category according to its level of political contestation. We estimate contestation using bill passage rates in the U.S. Congress and the Korean National Assembly over the past ten years, based on official legislative records~\citep{congressgov, nationalassemblybill}. Prior work treats legislative gridlock as an indicator of political conflict~\citep{Binder_1999, jones2001party, agarwal2024how}, making bill passage rates a useful proxy for issue-level contestation. 

For each country, we map bills to the eight issue categories and compute category-level passage rates as the proportion of introduced bills that are passed. We then allocate instances inversely proportional to passage rates, so that more contested categories receive more instances. Appendix~\ref{app:bill_passage_rate} reports the resulting category distribution and bill-to-category mapping procedure.

\paragraph{Opposing-position pairing.}



From the collected materials, we pair statements that take clear opposing positions on the same or closely related political issue along the relevant ideological axis. For example, Figure~\ref{fig:polar_process} shows two labor-related statements: one supports stronger labor rights, while the other argues for reforming labor laws to reduce restrictions. The pair addresses the same policy domain but represents opposite political positions.


We exclude statements that only criticize opponents, praise the speaker’s own party, or provide factual information without an explicit policy position. Within each category, we seek broad topic coverage while avoiding overrepresentation of a single policy debate, unless it appears through substantively different policy angles. Pairing is performed manually by the authors, with U.S. and South Korean instances reviewed by annotators familiar with the corresponding political contexts.

\paragraph{Instance construction.}
We convert each statement pair into an evaluation instance in three steps. First, we write a neutral context that introduces the issue without favoring either position. Second, we derive two political continuations from the paired source statements, preserving their substantive claims and rhetorical intensity while minimally paraphrasing them so that each continuation forms a complete sentence when joined with the context. Third, we add a semantically unrelated continuation that is grammatically compatible with the context but disconnected from the political issue. In Figure~\ref{fig:polar_process}, the matched pair is converted into an instance centered on a modern labor framework, with two opposed political continuations and one unrelated continuation.

To ensure that instances measure political positions rather than associations with specific political figures or contemporary events, we generalize overly specific references where necessary. For example, named politicians, party-specific slogans, branded policy labels, and numerical claims are replaced with more general descriptions or removed. We constrain instance length to roughly 200 characters for English and 100 characters for Korean.

\paragraph{Cross-lingual translation.}
To evaluate political bias across both context and language, each source instance is translated into the other language: U.S. instances are translated into Korean, and South Korean instances are translated into English. The final dataset therefore contains 4,026 instances, consisting of 2,013 original source instances and 2,013 translated instances. Translation is performed using GPT-5.5. Because subtle shifts in wording can undermine cross-lingual comparisons, all translated instances are audited for faithfulness to the source claim, preservation of rhetorical intensity, and fluency in the target language. Translations that do not meet these criteria are manually revised.

\paragraph{Review and revision.}

Each instance is drafted by one author and cross-checked by the other authors under a shared review guideline. Reviewers check the political neutrality of the context, connection between the context and political options, preservation of the source statements' meaning, grammatical correctness of all continuations, and semantic unrelatedness of the competence-check continuation. A context is considered neutral when it introduces the policy issue without favoring either political option or relying on explicit partisan or actor-specific cues. Based on the cross-checking logs, 10.3\% of U.S. instances and 13.4\% of South Korean instances required substantive revision or replacement before final inclusion, excluding minor wording or typo corrections. Flagged cases are discussed and revised, removed, or replaced as needed, followed by final adjudication by a bilingual author with expertise in law and public policy. The detailed review process is described in Appendix~\ref{app:review_instances}.

As a final validation step, we conduct an external review on a sampled subset of each dataset with ten reviewers. The external review focuses on grammaticality, fidelity to the original source pair, and other minor errors. Concrete feedback from the reviewers is also incorporated into the final dataset by replacing source pairs, correcting grammatical errors, or revising instance wording where necessary. Detailed procedures and results are provided in Appendix~\ref{app:external_review}.


%% file: latex/5_experiments.tex
\raggedbottom
\section{Experiments}
\label{sec:5_experiments}

\subsection{Experimental Setup}

\paragraph{Models.}

We evaluate 38 LLMs across both globally deployed and Korean-specialized model families. Global models include  Qwen3~\citep{yang2025qwen3}, Llama 3~\citep{grattafiori2024llama}, and Mistral~\citep{Jiang2023Mistral7}; Korean-specialized models include Kanana 1.5~\citep{bak2025kanana}, EXAONE-4.0~\citep{bae2025exaone}, A.X-4.0~\citep{skt2025ax40}, HyperCLOVAX~\citep{team2025hyperclova}, Solar~\citep{kim-etal-2024-solar}, Mi:dm~\citep{shin2026mi}, and Ko-GPT-Trinity~\citep{skt2025kogpttrinity}. This selection enables comparisons across model scales, developers, and language specializations. Full model details are provided in Appendix~\ref{app:model_details}.

\begin{figure*}[t]
    \centering
    \includegraphics[width=\textwidth]{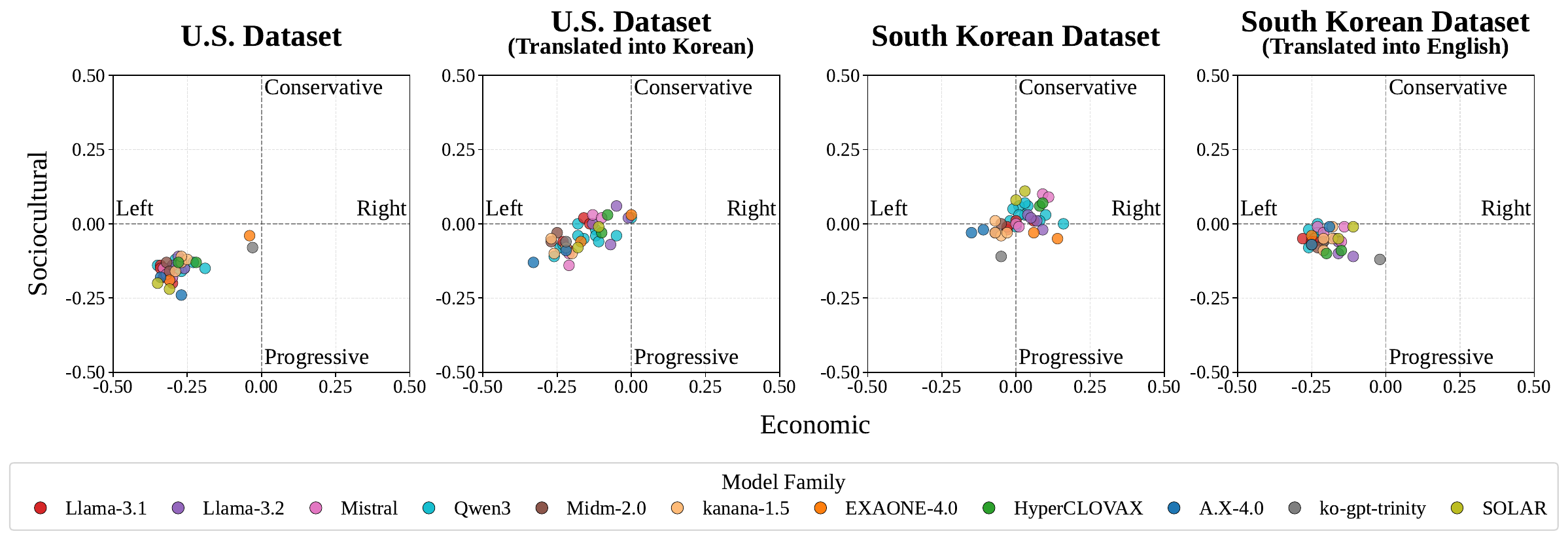}
    \caption{Political positions of LLMs on the economic and sociocultural axes. Each point represents a model, with colors indicating its model family. The displayed range is set to \([-0.5, 0.5]\) for readability, while the full position range is \([-1, 1]\). Appendix~\ref{app:political_position} provides the full-scale plot.}
    \label{fig:position_05}
    \vspace{-1.1em}    
\end{figure*}

\paragraph{Evaluation metrics.}

{
\setlength{\abovedisplayskip}{4pt}
\setlength{\belowdisplayskip}{3pt}
\setlength{\abovedisplayshortskip}{2pt}
\setlength{\belowdisplayshortskip}{2pt}

We evaluate models with the LM Evaluation Harness~\citep{biderman2024lessons}. Each candidate continuation is scored by length-normalized log-likelihood, and the highest-scoring option is treated as the model’s choice. Option 1 corresponds to the left position on the economic axis and the progressive position on the sociocultural axis, option 2 to the right/conservative position, and option 3 to a semantically unrelated option.

Polar reports political position and the ICAT (Idealized CAT) score~\citep{nadeem-etal-2021-stereoset}. The political position measures which of the two political options the model prefers within an axis or category. For each instance, we compare the length-normalized log-likelihoods of option 1 and option 2. We define \(p_1\) and \(p_2\) as the proportions of instances where option 1 and option 2 receive the higher score, respectively (e.g., \(p_1 = 0.6\), \(p_2 = 0.4\)). The position is then defined as:
\[
\mathrm{Position} = p_2 - p_1
\]

This comparison excludes option 3 and focuses only on the relative preference between the two political options. The position score ranges from \(-1\) to \(1\). A value close to \(-1\) indicates a left or progressive preference, while a value close to \(1\) indicates a right or conservative preference. For example, if option 1 (left) receives a higher likelihood than option 2 (right) in \(60\%\) of economic-axis instances, and option 2 receives a higher likelihood in \(40\%\), then \(p_1 = 0.6\) and \(p_2 = 0.4\). The resulting position score is \(p_2 - p_1 = -0.2\).


ICAT combines language modeling competence and ideological balance. Let \(\mathrm{LMS}\) denote the language modeling score and \(\mathrm{NS}\) denote the neutrality score. Following StereoSet, ICAT is defined as:
\[
\mathrm{ICAT} = \mathrm{LMS} \times \mathrm{NS}
\]
The language modeling score measures whether the model assigns higher likelihood to politically meaningful options than to the unrelated option. Let \(c_i\) denote the model's choice for instance \(i\). \(\mathrm{LMS}\) is defined as the percentage of instances where option 3 is not selected:
\[
\mathrm{LMS}
=
100 \times \frac{1}{N}\sum_{i=1}^{N}\mathbf{1}[c_i \neq 3]
\]
For example, if option 3 is selected in 20\% of instances, \(\mathrm{LMS}\) is 80.

The neutrality score captures how close the model's position is to zero:
\[
\mathrm{NS} = 1 - |\mathrm{Position}|
\]
Higher \(\mathrm{NS}\) indicates greater balance between two opposed options. We compute NS at the category-level and average across categories within each axis to prevent opposite directional biases from canceling out. We then average the economic and sociocultural ICAT scores to obtain the final score.

}

\subsection{Results}



\paragraph{Overall political position across contexts.}

We map each model to a two-dimensional political space using its economic and sociocultural position scores. Figure~\ref{fig:position_05} shows markedly different distributions across the two datasets. On the U.S. dataset, all models fall in the negative region on both axes, with economic scores ranging from \(-0.35\) to \(-0.03\) and sociocultural scores ranging from \(-0.24\) to \(-0.04\). In contrast, on the South Korean dataset, models cluster near the origin and appear on both sides of each axis, with economic scores ranging from \(-0.15\) to \(0.16\) and sociocultural scores ranging from \(-0.11\) to \(0.11\). Detailed model-level results are provided in Appendix~\ref{app:model_level_metrics}.

These results show a consistent left-progressive tendency across all evaluated models on the U.S. dataset, while model positions are more centered and mixed on the South Korean dataset. This pattern is consistent with prior work reporting left-leaning tendencies in LLMs, particularly in U.S. and Western European settings~\citep{bernardelle2025mapping, becchetti2026political, shu2026how}. One possible interpretation is that pretraining corpora or alignment procedures may increase the likelihood of policy language associated with particular ideological or demographic groups~\citep{rozado2023danger, santurkar2023whose, fulay-etal-2024-relationship}. The contrast between the two datasets indicates that political bias should be assessed across multiple political contexts, as model preferences identified in one context may not generalize to others.


\paragraph{Category-level bias patterns.}

Category-level results reveal political preferences that are not visible in aggregate axis-level scores. In the U.S. dataset, category-level mean scores are left or progressive leaning across all eight issue categories, although their strength varies by domain. \textit{Welfare State} and \textit{Gender / Minorities / Equality} show the strongest left-progressive tendencies, with mean position scores of \(-0.42\) and \(-0.33\). In contrast, \textit{International Relations} and \textit{National Defense / Security} are closer to the origin, with mean scores of \(-0.08\) and \(-0.03\). Detailed plots and results are provided in Appendix~\ref{app:category_level_position} and Appendix~\ref{app:category_level_metrics}.

These results suggest that even when the overall direction of political preference is consistent, its strength depends on the issue being evaluated. One possible interpretation is that left-progressive expressions in topics such as \textit{healthcare access} and \textit{minority protection} overlap with safety- and inclusion-oriented language frequently emphasized during instruction tuning and alignment~\citep{ouyang2022training}. By contrast, categories such as \textit{International Relations} may contain policy language that is less strongly associated with a single ideological direction in training or alignment data.

The South Korean dataset shows a more mixed pattern. Most categories remain close to the origin, but several categories exhibit clear directional tendencies. For example, \textit{Market Economy} is right-leaning, with a mean score of \(0.18\), while \textit{Labor} is left-leaning, with a mean score of \(-0.16\). These opposing category-level tendencies can partially cancel out at the axis level, making models appear more neutral in aggregate than they are within specific issue domains.



\paragraph{Korean-specialized vs global models.}

We compare Korean-specialized and globally deployed models to examine how model group differences vary across political contexts. We quantify these differences using the average category-level gap, defined as the mean absolute difference between the two groups' position scores across categories.

On the U.S. dataset, the two groups show broadly similar category-level patterns. Both exhibit stronger progressive tendencies in \textit{Welfare State} and \textit{Gender / Minorities / Equality} and remain closer to the origin in \textit{International Relations} and \textit{National Defense / Security}. The average category-level gap is \(0.03\), suggesting that Korean-specialized and globally deployed models produce similar patterns in U.S. political content. 

On the South Korean dataset, the gap increases to \(0.10\). The largest gaps appear in \textit{Market Economy} and \textit{Trade / Energy}, where Korean-specialized models score \(0.08\) and \(-0.07\) respectively, while global models score \(0.25\) and \(0.07\).

These results indicate that Korean-specialized and global models are more similar on U.S. political content than on South Korean political content. One possible explanation is that South Korean political expressions reflect local issue framing, media discourse, and policy language, which may be captured differently by models trained more heavily for Korean language usage. Many Korean-specialized models are built on open-weight global base models, which may explain why they retain similar tendencies to global models on U.S. content~\citep{kim-etal-2024-solar, skt2025ax40}.


\begin{figure}[t]
    \centering
    \includegraphics[width=1\linewidth]{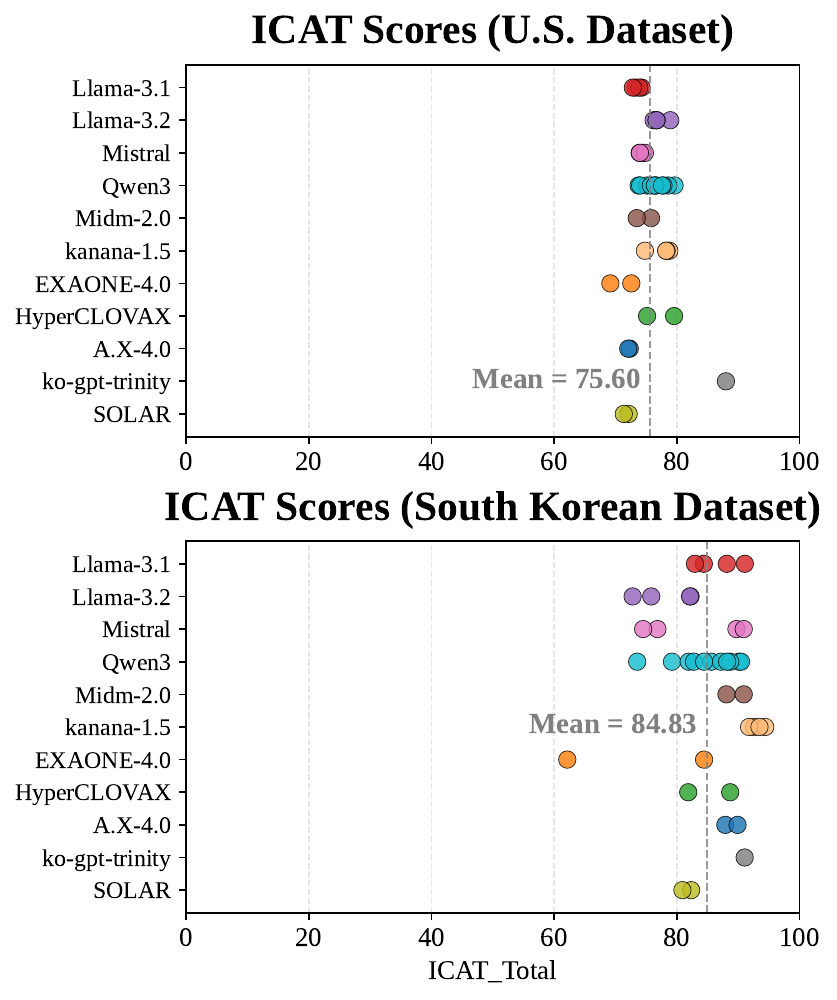}
    \vspace{-0.6em}
    \caption{ICAT scores across models on the two datasets. Each point represents an individual model belonging to the corresponding model family.} 
    \label{fig:icat_distribution}
    \vspace{-1.2em}
\end{figure}

\paragraph{Comparison across global model families.}
We further compare Qwen3, Mistral, and Llama 3 to examine whether globally deployed model families show distinct political tendencies. On the original U.S. dataset, the economic and sociocultural positions are (-0.28, -0.15) for Qwen3, (-0.32, -0.17) for Mistral, and (-0.30, -0.15) for Llama 3. On the South Korean dataset, the corresponding positions remain close to the origin: (0.04, 0.02), (0.05, 0.04), and (0.02, -0.01), respectively. In comparison, presentation-language shifts are larger and consistent across the three families. When U.S. content is translated into Korean, the economic and sociocultural positions shift by (+0.13, +0.10) for Qwen3, (+0.16, +0.12) for Mistral, and (+0.18, +0.14) for Llama 3; when South Korean content is translated into English, they shift by (-0.28, -0.06), (-0.23, -0.07), and (-0.22, -0.06), respectively. These results indicate that differences across the three model families are relatively small, whereas presentation language has a consistent effect.

\paragraph{ICAT scores.}



We examine ICAT scores, which combine ideological balance with language modeling competence. A high ICAT score requires both low directional political preference and high LMS. Comparable likelihoods for the two political options count as meaningful balance only when the model also ranks them above the unrelated option. 

Most models achieve high LMS, indicating that they generally distinguish politically relevant continuations from unrelated continuations. EXAONE-4.0-1.2B is a notable exception, with average LMS scores of \(75.68\) and \(75.51\) for each dataset, suggesting weaker language modeling competence in this evaluation setting. 

ICAT scores differ substantially between the two datasets. The mean ICAT score is \(84.83\) on the South Korean dataset, compared with \(75.60\) on the U.S. dataset. This difference mainly reflects the consistent left-progressive preferences on the U.S. dataset, which reduce the neutrality component even when models maintain high LMS. Thus, ICAT complements position scores by indicating whether low directional preference is supported by adequate language-modeling competence.



\paragraph{Effect of presentation language.}

We define presentation language as the language in which the same political content is evaluated, independent of the political context from which the content was originally drawn. To examine its effect, we compare original and translated Polar instances: U.S. political content in English and Korean, and South Korean content in Korean and English.

Figure~\ref{fig:position_05} shows that presentation language substantially affects measured political preferences. When South Korean political content is presented in English, model distributions shift from near the origin toward the left-progressive direction. Conversely, when U.S. political content is presented in Korean, the originally left-progressive distribution moves closer to the origin. The shift is more pronounced on the economic axis than on the sociocultural axis. 

The results indicate that measured bias is influenced not only by political content but also by the language of presentation. A plausible explanation is that the presentation language alters the linguistic and cultural cues models use to evaluate political content, leading to the same underlying content receiving different relative likelihoods in English and Korean presentations~\citep{li2024culturellm, xu2025survey}.




%% file: latex/6_conclusion.tex
\raggedbottom
\section{Conclusion}
\label{sec:6_conclusion}

This work introduces Polar, a two-axis, eight-category benchmark designed to evaluate political bias in LLMs across U.S. and South Korean political contexts. By leveraging option-level likelihoods, Polar measures directional political preferences while concurrently assessing language modeling competence. An analysis of 38 LLMs reveals consistent leftward progressiveness in choices on U.S. political content, whereas choices on South Korean content are more centered, mixed, and dependent on specific categories. Language translation experiments indicate that the language of presentation affects measured bias, suggesting that models may assign different likelihoods to identical political content depending on linguistic and cultural framing. These findings highlight the need for multilingual, cross-contextual evaluation to achieve a comprehensive understanding of political bias in LLMs.


%% file: latex/limitations.tex
\section*{Limitations}


Polar has three main limitations. First, it reflects U.S. and South Korean political contexts from 2016 to 2025. Since political coalitions, issue salience, and public debates change over time, future work should update the dataset to capture emerging issues and shifts in party positions. Second, Polar covers only two countries. Extending it to additional countries, languages, and cultural settings would support broader multilingual and cross-cultural evaluation. Third, our approach requires access to token-level likelihoods. As a result, the method is limited to open-weight models for which these scores can be computed, but not to closed-source systems that expose only generated outputs. The evaluated model set should therefore not be interpreted as representative of all current frontier LLMs. We seek to develop complementary evaluation methods for such models as future work.

%% file: latex/ethical_considerations.tex
\section*{Ethical Considerations}

This work evaluates political bias in LLMs, a politically and socially sensitive topic. Our goal is not to endorse or oppose any party, ideology, or policy position, but to measure relative model preferences between politically opposed options under a controlled setting. Results should be interpreted as likelihood-based preferences under Polar, not as evidence that models possess political beliefs, intentions, or stable ideological commitments.

Political bias benchmarks may be misused if results are overgeneralized or treated as fixed model properties. They may also unfairly label particular models, languages, or communities. To reduce this risk, we report bias as relative preference within specific political contexts, languages, and issue categories. Our findings show that measured bias varies across context, category, model group, and presentation language, cautioning against broad conclusions from a single benchmark.

Polar does not contain private user data, and our experiments do not involve human users. The benchmark is constructed from public political materials rewritten into neutral contexts and politically opposed continuations. We release Polar for reproducible, diagnostic, and comparative analysis of political bias in LLMs, not for ranking models as politically acceptable or unacceptable.

%% file: latex/acknowledgments.tex
\section*{Acknowledgments}
This work was partially supported by the National Research Foundation of Korea (NRF) under Grant No. RS-2023-00222663 (Center for Optimizing Hyperscale AI Models and Platforms) and under Grant No. 2023S1A5A8076585 (Early Career Researchers, Humanities and Social Sciences), and by the Institute for Information and Communications Technology Promotion (IITP) under Grant No. 2018-0-00581 (CUDA Programming Environment for FPGA Clusters) and No. RS-2025-02304554 (Efficient and Scalable Framework for AI Heterogeneous Cluster Systems), all funded by the Ministry of Science and ICT (MSIT) of Korea. It was also partially supported by the Korea Health Industry Development Institute (KHIDI) under Grant No. RS-2025-25454559 (Frailty Risk Assessment and Intervention Leveraging Multimodal Intelligence for Networked Deployment in Community Care), funded by the Ministry of Health and Welfare (MOHW) of Korea. Additional support was provided by the BK21 Plus Program for Innovative Data Science Talent Education (Department of Data Science, Seoul National University, No. 5199990914569) and the BK21 FOUR Program for Intelligent Computing (Department of Computer Science and Engineering, Seoul National University, No. 4199990214639), both funded by the Ministry of Education (MOE) of Korea. This work was also partially supported by the Advanced GPU Utilization Support Program, funded by the Ministry of Science and ICT (MSIT) of Korea and operated by the National IT Industry Promotion Agency (NIPA). Research facilities were provided by the Institute of Computer Technology (ICT) at Seoul National University. 


%% file: latex/appendix.tex
\appendix
\raggedbottom

\section{Details of Polar Taxonomy}
\label{app:polar_taxonomy}

\input{tables/appendix/polar_taxonomy}

We provide the full mapping between the Polar categories and the Manifesto Project coding scheme. In the Manifesto Project Dataset, each political statement is annotated with a Manifesto code. Each code represents a policy subcategory and is associated with an ideological direction. We use these code-level annotations to ground the category and direction labels in Polar. Table~\ref{tab:polar_taxonomy} lists the economic and sociocultural categories in Polar, their directional labels, and the corresponding Manifesto codes and subcategories.

This mapping serves two purposes. First, it ensures that each Polar category is grounded in an established coding scheme rather than in ad-hoc ideological labels. Second, it provides a consistent basis for grouping Manifesto-annotated statements into Polar categories. For supplementary sources that do not come with Manifesto annotations, such as party documents, press releases, and news articles, we apply the same Manifesto codebook criteria to determine the corresponding category and direction.

\begin{figure}[t]
    \centering
    \includegraphics[width=\linewidth]{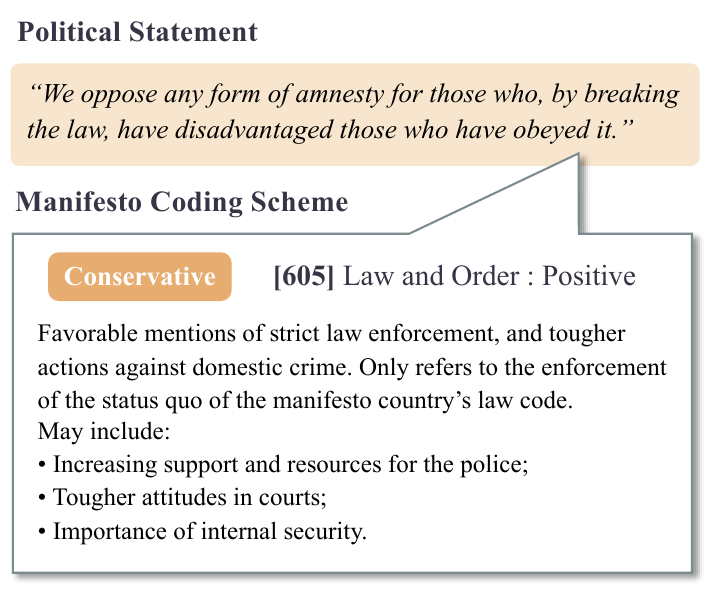}
    \caption{Example of a political statement annotated with a Manifesto policy code and its corresponding political direction.}
    \label{fig:manifesto_code_example}
\end{figure}

Figure~\ref{fig:manifesto_code_example} shows an example of a Manifesto-annotated political statement and how its existing code is linked to Polar's taxonomy. The statement is labeled in the Manifesto Project Dataset with code 605, \textit{Law and Order: Positive}. This code refers to favorable mentions of strict law enforcement and tougher responses to crime, and it corresponds to the conservative direction on the sociocultural axis in Polar.

\section{Category-Level Bill Passage Rates}
\label{app:bill_passage_rate}
\input{tables/appendix/bill_passage_rate}

We use official legislative records to estimate the level of political contestation in each category. For each country, we collect bills introduced over the past ten years, from 2016 to 2025, and compute the passage rate of bills corresponding to each Polar category. For the U.S. dataset, we use records from the~\citet{congressgov}. Since each bill is annotated with a policy area, such as \textit{Taxation} or \textit{Minority Issues}, we map these policy-area labels to the Polar categories. For the South Korean dataset, we use records from~\citet{nationalassemblybill}. Since South Korean bill records are organized by the standing committee that proposes or reviews each bill, we map standing committees to the Polar categories.

Table~\ref{tab:bill_passage_allocation} reports the bill passage rate for each category and the corresponding number of instances allocated in Polar. As described in the main text, we assign more instances to categories with lower passage rates, treating lower passage rates as evidence of stronger political contestation.

\section{Details of the Dataset Construction Process}
\label{app:dataset_construction_details}

In this section, we provide additional details about the construction and review procedure for Polar. While the main text describes the overall pipeline, we focus here on how candidate statements are selected, how complete instances are reviewed, and how final instances are revised and selected. The procedure consists of four stages: (1) preparing candidate instances, (2) reviewing complete instances, (3) revising and selecting final instances, and (4) conducting external validation.

\subsection{Constructing Candidate Instances}
\label{app:preparing_instances_review}

Before review, each author constructs candidate instances for assigned categories. For each category, the author examines statements associated with Manifesto codes mapped to the corresponding Polar subcategories and lists candidate statements for both sides of a political issue. We prioritize statements that express a clear political claim and exclude statements that are too general, too simple, or difficult to convert into evaluation options.

The author then matches statements that address the same or closely related issue but express opposing positions. A source statement can be reused when necessary, but we aim to cover diverse issues and resources. When the same source is used more than once, we focus on different claims or policy aspects so that the resulting instances do not repeatedly test the same narrow issue.

Each matched pair is converted into a complete evaluation instance. The author writes a neutral context, derives two political continuations from the paired statements, and adds a semantically unrelated continuation. The unrelated option is written to be grammatical as a continuation but semantically disconnected from the political issue. We use non-political topics, such as daily life, transportation, nature, food, sensory experience, and ordinary actions, to avoid unintended political associations.

\subsection{Reviewing Instances}
\label{app:review_instances}

Each complete instance is cross-checked by the other authors under the shared review guideline. Reviewers examine the context, the two political options, and the unrelated option together. The review focuses on the following criteria:

\begin{itemize} [noitemsep, topsep=0pt]
    \item whether the context is politically neutral and does not favor either political option;
    \item whether the context and each political option form a grammatical and semantically natural continuation;
    \item whether the political options preserve the claim and rhetorical intensity of the original paired statements;
    \item whether the paired statements and the resulting options represent opposing positions on the same issue, based on the Manifesto coding scheme;
    \item whether the unrelated option is grammatical as a continuation while remaining semantically unrelated to the political issue;
    \item whether the unrelated option avoids keywords or expressions that could evoke the political issue in the context;
    \item whether the instance contains typographical errors or other surface-level mistakes.
\end{itemize}

For context neutrality, reviewers check whether the context introduces the policy issue without favoring either political option. We avoid explicit references to party names, named politicians, directional modifiers, or evaluative claims that could directly signal one political position. This is intended to ensure that model preferences reflect the policy content of the continuations rather than partisan or actor-specific cues.

Reviewers directly revise issues that can be resolved through wording changes, grammatical correction, or typo correction. When an instance raises a more substantive issue, such as unclear ideological contrast, weak source preservation, or an ambiguous unrelated option, the authors discuss whether to revise the instance, replace the original pair, or remove it and construct a new instance.

\subsection{Revising and Selecting Final Instances}
\label{app:selecting_instances}

After the first review, we revise, remove, or replace instances according to the reviewers' feedback. We then conduct an additional pass over the revised instances and select the final instances to include in the dataset.

When the number of constructed instances exceeds the target allocation based on bill passage rates, we reduce redundancy within each category. In particular, we limit repeated coverage of the same issue so that the final dataset covers a broader range of political topics. We prioritize instances that better satisfy the review criteria and provide clearer, more balanced representations of opposing political positions.

For translated instances, we additionally check whether the translation preserves the source claim, rhetorical intensity, and grammatical structure of the original instance. Translations that introduce meaning shifts, weaken the political contrast, or create unnatural continuations are manually revised.

\subsection{External Validation}
\label{app:external_review}

As the final stage of the review process, we conduct an external validation on approximately half of each source dataset. We sample 501 of the 1,004 U.S. instances and 506 of the 1,009 South Korean instances. 10 external reviewers examine the sampled instances using three criteria: grammaticality, source fidelity, and error checking.

The external reviewers were recruited from members of the same university with diverse academic backgrounds and disciplines, including psychology, sociology, engineering, and data science. Reviewers were selected based on their proficiency in the target language (English or Korean) and sufficient familiarity with the political context and issues of the corresponding country (the United States or South Korea). Before participating, reviewers were informed of the purpose of the study, the external validation procedure, and how their review results would be used. Detailed information provided to the reviewers, including the review procedure, potential risks, confidentiality, and compensation, is reported in Appendix~\ref{app:participant_information}.

Grammaticality evaluates whether the context forms a grammatically correct continuation with each of the three options. Source fidelity evaluates whether options 1 and 2 preserve the content and substantive claim of the corresponding source pair. Error checking identifies typographical errors and other issues not covered by the first two criteria.

The resulting inter-annotator agreement rate is 88.02\% for the U.S. dataset and 89.13\% for the South Korean dataset. Among the reported issues in the U.S. dataset, 40.8\% concern grammaticality, 31.6\% source fidelity, and 27.6\% typographical or other errors. In the South Korean dataset, source fidelity accounts for 53.8\% of the reported issues, followed by grammaticality at 38.5\% and typographical or other errors at 7.7\%.

We review all feedback from the external reviewers and revise the corresponding instances before finalizing the dataset. Depending on the reported issue, revisions include replacing the original source pair with a more appropriate pair, correcting grammatical errors, or revising the wording of the context or options. These revisions are incorporated into the final version of Polar. Together, the author review and external validation processes improve the consistency and quality of the dataset before final inclusion.

\subsection{Information for Research Participants}
\label{app:participant_information}

\textbf{Research Title:} Benchmark Development for Evaluating Political Bias in Large Language Models and Analysis of Model Bias Tendencies

\textbf{Principal Investigators:} Jaejin Lee (Professor, Graduate School of Data Science, Seoul National University), Sangho Kim (Master's Student, Graduate School of Data Science, Seoul National University)

This study aims to validate the data quality of a benchmark developed to evaluate political bias in large language models (LLMs). Polar is a multiple-choice benchmark designed to assess LLM preferences between opposing political positions in the political contexts of the United States and South Korea.

In this study, external reviewers examine a subset of the instances created during the benchmark construction process and assess grammaticality, whether the content and claims of the original source materials are appropriately reflected, and whether there are typographical or other errors. Through this process, we aim to further validate the quality of the dataset from the perspective of reviewers who are independent of the authors. You have been invited to participate in this study because you are an adult who is familiar with the language and political context of the corresponding dataset.

This study will be conducted remotely using Google Spreadsheet. After reviewing the task guidelines provided by the research team, participants will review the assigned data according to predefined criteria.

Participation in this study is entirely voluntary. Before deciding whether to participate, it is important that you understand why this study is being conducted and what tasks you will be asked to perform. Please read the following information carefully before deciding whether to participate. If you have any questions about the study, you may contact the research team.

\begin{enumerate}
    \item \textbf{Why is this study being conducted?}

    This study is conducted to validate, through external reviewers, the data quality of Polar, a benchmark for evaluating political bias in large language models.

    Polar is designed to evaluate whether a language model relatively prefers one of two opposing political positions on politically contested issues. It covers the political contexts of the United States and South Korea and addresses a range of policy and political issues along the Economic and Sociocultural axes.

    In this study, external reviewers who are not authors of the dataset independently examine a subset of the data to assess the grammatical appropriateness of each instance, consistency with the original source materials, and the presence of typographical or other errors. The purpose is to validate the reliability and quality of the dataset.

    \item \textbf{How many people will participate?}

    A total of 10 external reviewers are expected to participate in this study. Participants are recruited from adults who meet the requirements of fluency in the relevant language (Korean or English) and familiarity with the political context of the corresponding country.

    \item \textbf{What will happen if I participate in the study?}

    If you agree to participate, you will first receive an explanation of the Polar dataset and the external review task, along with task guidelines from the research team.

    You will then review approximately 500 instances through Google Spreadsheet. Each instance contains original source materials related to political or social issues, as well as a context and options constructed on the basis of those materials.

    For each instance, you will review the following three criteria:

    \begin{itemize}
        \item \textbf{Grammaticality:} Determine whether there are any grammatically incorrect or unnatural parts when the context is combined with each option.
        \item \textbf{Source Fidelity:} Determine whether the content and claims of the original source data (pair) are appropriately reflected in the corresponding option.
        \item \textbf{Error Check:} Determine whether the instance contains any typographical errors or other clear errors.
    \end{itemize}

    If you identify a problem, you will select the corresponding issue type and provide details about the issue. During the review process, you will not directly modify the dataset. You will only record comments on parts that you consider problematic.

    This study is not intended to investigate participants' political orientation or whether they agree with any particular political position. Participants are not asked to express personal preferences regarding the political claims presented. Instead, they are asked to assess the quality of the dataset instances according to the criteria described above.

    \item \textbf{How long will participation take?}

    You will review approximately 500 data instances, and the entire task is expected to take approximately 4--5 hours. Reviewing an individual instance may take approximately 30--40 seconds on average. The actual time required may vary depending on the individual reviewer's pace. The review task will be conducted from September 8 (Tuesday) to September 11 (Friday).

    \item \textbf{Can I withdraw from the study after participation has begun?}

    Yes. You may discontinue participation at any time without any disadvantage or penalty. If you wish to withdraw from the study, please inform a member of the research team or one of the principal investigators. If you withdraw, any review data you have already provided will be immediately discarded.

    \item \textbf{Are there any side effects or risks?}

    There are almost no anticipated physical risks associated with participation in this study. However, because the data reviewed in this study include politically and socially contested policies and issues, you may experience discomfort while reviewing some instances.

    In addition, reviewing a large number of instances over approximately 3--4 hours may cause fatigue. Participants may freely take breaks whenever necessary. If you experience discomfort or any other problem during the review process, please inform the research team.

    \item \textbf{Are there any benefits to participating in this study?}

    There are no direct benefits to you from participating in this study. However, the review results you provide may be used to validate and improve the quality of the Polar dataset for evaluating political bias in large language models. In particular, your independent assessment as an external reviewer will contribute to evaluating the grammatical appropriateness of the dataset, its fidelity to the original source materials, and the presence of other errors.

    \item \textbf{Will there be any disadvantage if I choose not to participate?}

    You are free to decline participation in this study. There will be no disadvantage or penalty if you choose not to participate or decide to withdraw after participation has begun. If you are a student, your decision regarding participation will have no relationship to your classes, grades, degree program, or any other academic evaluation.

    \item \textbf{Will the confidentiality of all personal information collected in this study be protected?}

    In this study, the review results that participants provide for individual data instances will be collected as research data. Anonymous IDs will be used for the review results. When the results of this study are used in academic papers, presentations, or other research outputs, no personally identifiable information about participants will be disclosed.

    \item \textbf{Will I receive compensation for participating in this study?}

    If you complete the assigned review task, you will receive a gift voucher worth KRW 50,000 as compensation for your participation. Delivery of the compensation may take some time due to administrative procedures related to research funding. However, if you discontinue participation before completing the task, compensation cannot be provided.

    For the purpose of providing compensation, the researchers may collect personal information including your name, date of birth, affiliation, and contact information. If such personal information is collected for compensation purposes, it will be managed separately from the research data and destroyed immediately after the compensation has been provided.

    \item \textbf{Who should I contact if I have questions about the study?}

    If you have any questions about this study or experience any problems during participation, please contact the following member of the research team.

    Name: Sangho Kim \\
    Phone: 010-2711-4931 \\
    Email: \texttt{ksh4931@snu.ac.kr}
\end{enumerate}

\section{Details of the Models}
\label{app:model_details}
A total of 38 LLMs are evaluated in this paper. We select models from diverse developers and model scales to examine political bias across model families and deployment targets. We group the individual models into two broad model groups: global and Korean-specialized. The global group consists of models from broadly used international model families, while the Korean-specialized group consists of models developed with an explicit focus on Korean language use or South Korean user contexts.

\subsection*{Global Models}
\begin{itemize}
\small

\item{\textbf{Llama 3.1 and Llama 3.2}}
\begin{itemize}[leftmargin=6mm,itemsep=0pt,parsep=0pt,topsep=0pt]
    \item{{Llama-3.1-8B}}
    \item{{Llama-3.1-8B-instruct}}
    \item{{Llama-3.1-70B}}
    \item{{Llama-3.1-70B-instruct}}
    \item{{Llama-3.2-1B}}
    \item{{Llama-3.2-1B-instruct}}
    \item{{Llama-3.2-3B}}
    \item{{Llama-3.2-3B-instruct}}
\end{itemize}

\item{\textbf{Mistral}}
\begin{itemize}[leftmargin=6mm,itemsep=0pt,parsep=0pt,topsep=0pt]
    \item{{Mistral-7B-v0.3}}
    \item{{Mistral-7B-Instruct-v0.3}}
    \item{{Mistral-Small-24B-Base-2501}}
    \item{{Mistral-Small-24B-Instruct-2501}}
\end{itemize}

\item{\textbf{Qwen3}}
\begin{itemize}[leftmargin=6mm,itemsep=0pt,parsep=0pt,topsep=0pt]
    \item{{Qwen3-0.6B-Base}}
    \item{{Qwen3-0.6B}}
    \item{{Qwen3-1.7B-Base}}
    \item{{Qwen3-1.7B}}
    \item{{Qwen3-4B-Base}}
    \item{{Qwen3-4B}}
    \item{{Qwen3-8B-Base}}
    \item{{Qwen3-8B}}
    \item{{Qwen3-14B-Base}}
    \item{{Qwen3-14B}}
    \item{{Qwen3-32B}}
\end{itemize}

\end{itemize}

\subsection*{Korean-specialized Models}
\begin{itemize}
\small

\item{\textbf{Midm-2.0}}
\begin{itemize}[leftmargin=6mm,itemsep=0pt,parsep=0pt,topsep=0pt]
    \item{{Midm-2.0-Mini-Instruct} (2.3B)}
    \item{{Midm-2.0-Base-Instruct} (11.5B)}
\end{itemize}

\item{\textbf{kanana-1.5}}
\begin{itemize}[leftmargin=6mm,itemsep=0pt,parsep=0pt,topsep=0pt]
    \item{{kanana-1.5-2.1b-base}}
    \item{{kanana-1.5-2.1b-instruct-2505}}
    \item{{kanana-1.5-8b-base}}
    \item{{kanana-1.5-8b-instruct-2505}}
\end{itemize}

\item{\textbf{EXAONE-4.0}}
\begin{itemize}[leftmargin=6mm,itemsep=0pt,parsep=0pt,topsep=0pt]
    \item{{EXAONE-4.0-1.2B}}
    \item{{EXAONE-4.0-32B}}
\end{itemize}

\item{\textbf{HyperCLOVA X SEED}}
\begin{itemize}[leftmargin=6mm,itemsep=0pt,parsep=0pt,topsep=0pt]
    \item{{HyperCLOVAX-SEED-Text-Instruct-0.5B}}
    \item{{HyperCLOVAX-SEED-Text-Instruct-1.5B}}
\end{itemize}

\item{\textbf{A.X-4.0}}
\begin{itemize}[leftmargin=6mm,itemsep=0pt,parsep=0pt,topsep=0pt]
    \item{{A.X-4.0-Light} (7B)}
    \item{{A.X-4.0} (72B)}
\end{itemize}

\item{\textbf{ko-gpt-trinity}}
\begin{itemize}[leftmargin=6mm,itemsep=0pt,parsep=0pt,topsep=0pt]
    \item{{ko-gpt-trinity-1.2B-v0.5}}
\end{itemize}

\item{\textbf{SOLAR}}
\begin{itemize}[leftmargin=6mm,itemsep=0pt,parsep=0pt,topsep=0pt]
    \item{{SOLAR-10.7B-v1.0}}
    \item{{SOLAR-10.7B-Instruct-v1.0}}
\end{itemize}

\end{itemize}

\section{Details of the Experimental Settings}
We conduct all experiments using the LM Evaluation Harness~\citep{biderman2024lessons}, a library for evaluating LLMs across diverse benchmarks. For each model, we load the open-source weights and tokenizer from the Hugging Face Hub~\citep{wolf2019huggingface} and evaluate the model in a multiple-choice setting. Given a context, the evaluation script computes the log-likelihood of each candidate option conditioned on that context.

We score each option using length-normalized log-likelihood. Since raw log-likelihood tends to decrease as token length increases, longer options can be penalized. We divide each option's log-likelihood by its token length and use the normalized score to determine the model's choice. We then use a separate analysis script to compute political position, LMS, NS, and ICAT, and to generate the plots reported in the paper. Since the evaluation scores fixed candidate continuations without stochastic decoding, the results are deterministic under the same model weights, tokenizer, and evaluation setting. We verified this by repeating the evaluation three times under the same setting and obtained identical results across all runs. All experiments are conducted on a server with eight NVIDIA GeForce RTX 3090 GPUs.

\paragraph{Environmental impact.} 
We estimate the environmental impact of our evaluation using the Green Algorithms calculator and our runtime logs. Evaluating all 38 models required 79.1 GPU-hours on eight NVIDIA GeForce RTX 3090 GPUs, corresponding to an estimated energy consumption of 50.25 kWh and carbon emissions of 21.24 kg CO2e.

\section{Evaluation Results}
\label{app:evaluation_results}

In this section, we provide plots and tables from experiments that support the main results in Section~\ref{sec:5_experiments}. We report model positions, category-level position distributions, comparisons between global and Korean-specialized model groups, and full metric tables.

\subsection{Political Position of LLMs}
\label{app:political_position}

\begin{figure*}[p]
    \centering
    \includegraphics[width=1\textwidth]{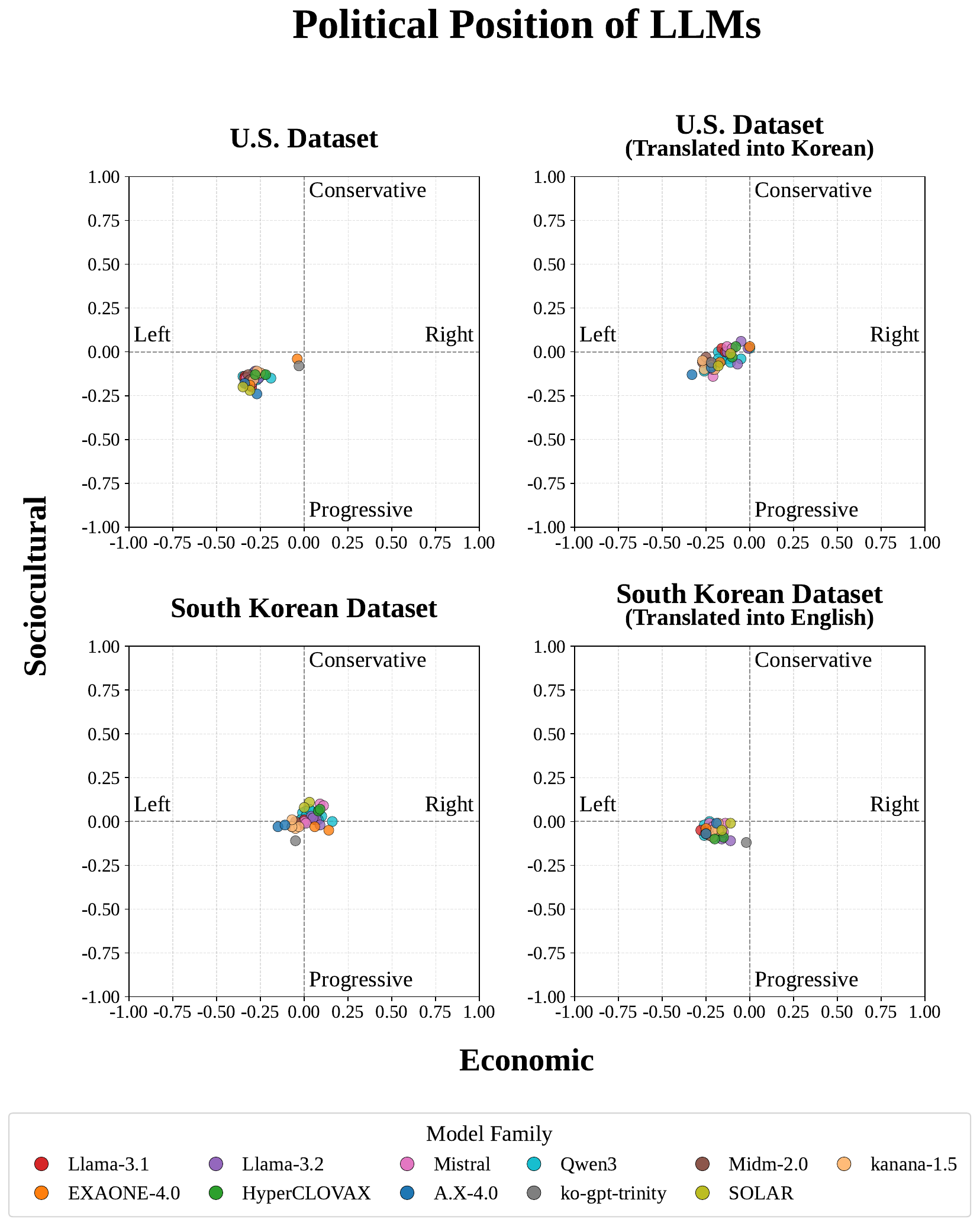}
    \caption{Two-dimensional political positions of LLMs across original and translated datasets. The x-axis represents the economic position, where negative values indicate left-leaning preferences and positive values indicate right-leaning preferences. The y-axis represents the sociocultural position, where negative values indicate progressive preferences and positive values indicate conservative preferences. Each point corresponds to a model.}
    \label{fig:political_position_1scale}
\end{figure*}

Figure~\ref{fig:political_position_1scale} shows the political positions of all evaluated models on the original and translated datasets. This figure uses the full \([-1,1]\) range for both axes. Each plot shows the positions of models for the dataset. The translated datasets further show that the same political content can lead to different position distributions depending on language form.

\subsection{Category-Level Position}
\label{app:category_level_position}

\begin{figure*}[p]
    \centering
    \includegraphics[width=1\textwidth]{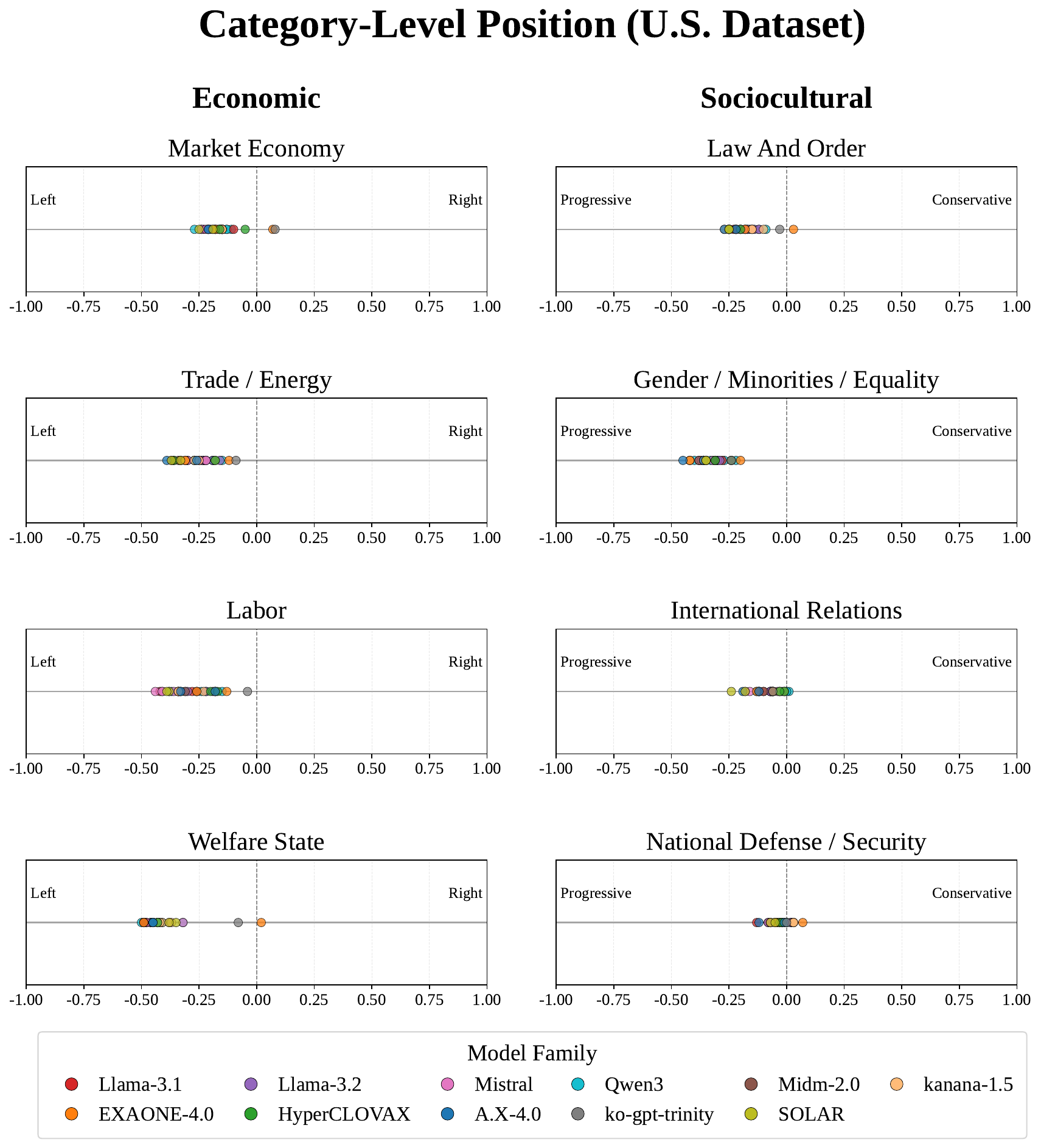}
    \caption{Category-level political positions on the U.S. dataset. The left column reports economic categories, and the right column reports sociocultural categories. Each point corresponds to a model.}
    \label{fig:us_category}
\end{figure*}

\begin{figure*}[p]
    \centering
    \includegraphics[width=1\textwidth]{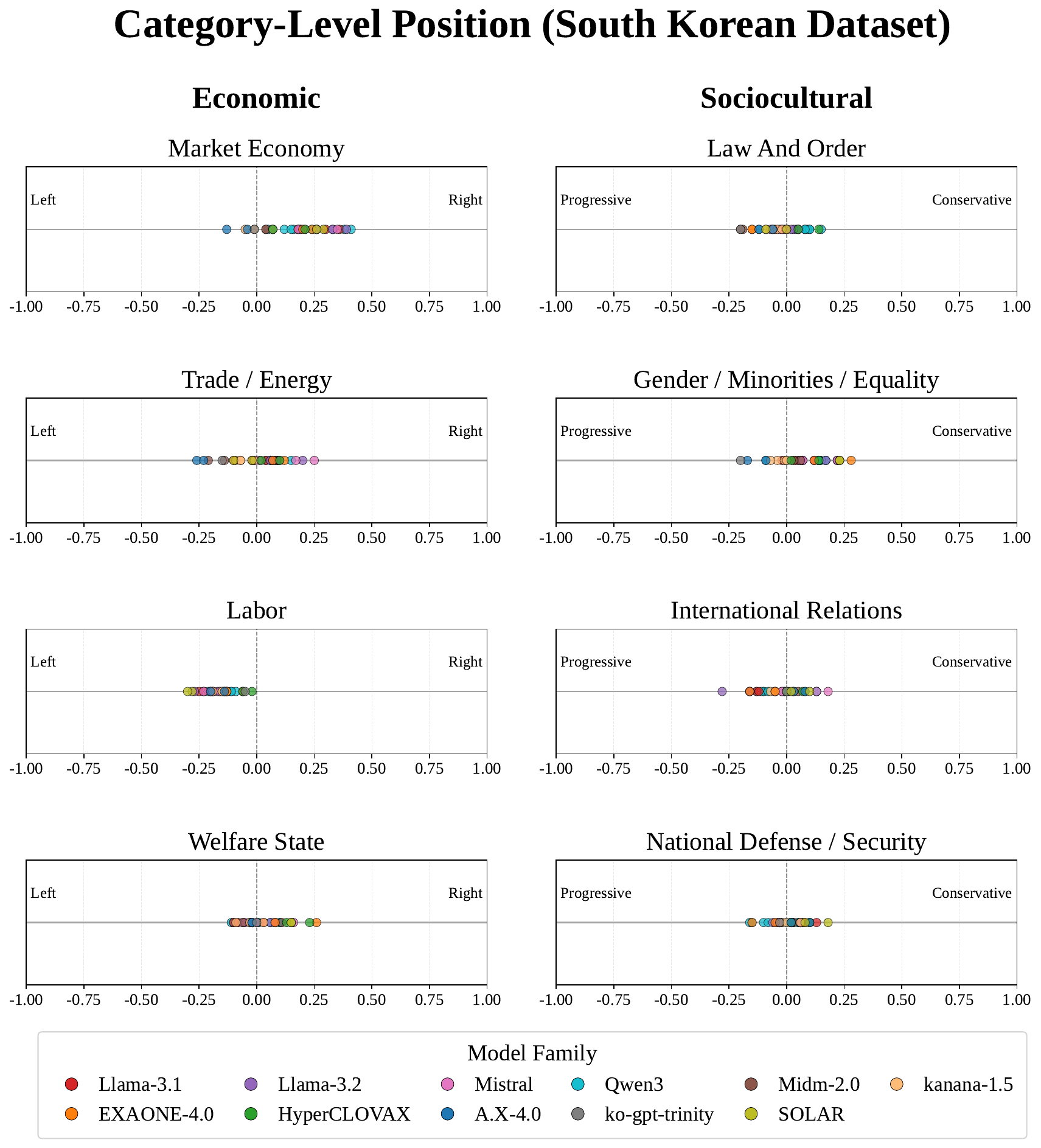}
    \caption{Category-level political positions on the South Korean dataset. The left column reports economic categories, and the right column reports sociocultural categories. Each point corresponds to a model.}
    \label{fig:ko_category}
\end{figure*}

\begin{figure*}[p]
    \centering
    \includegraphics[width=1\textwidth]{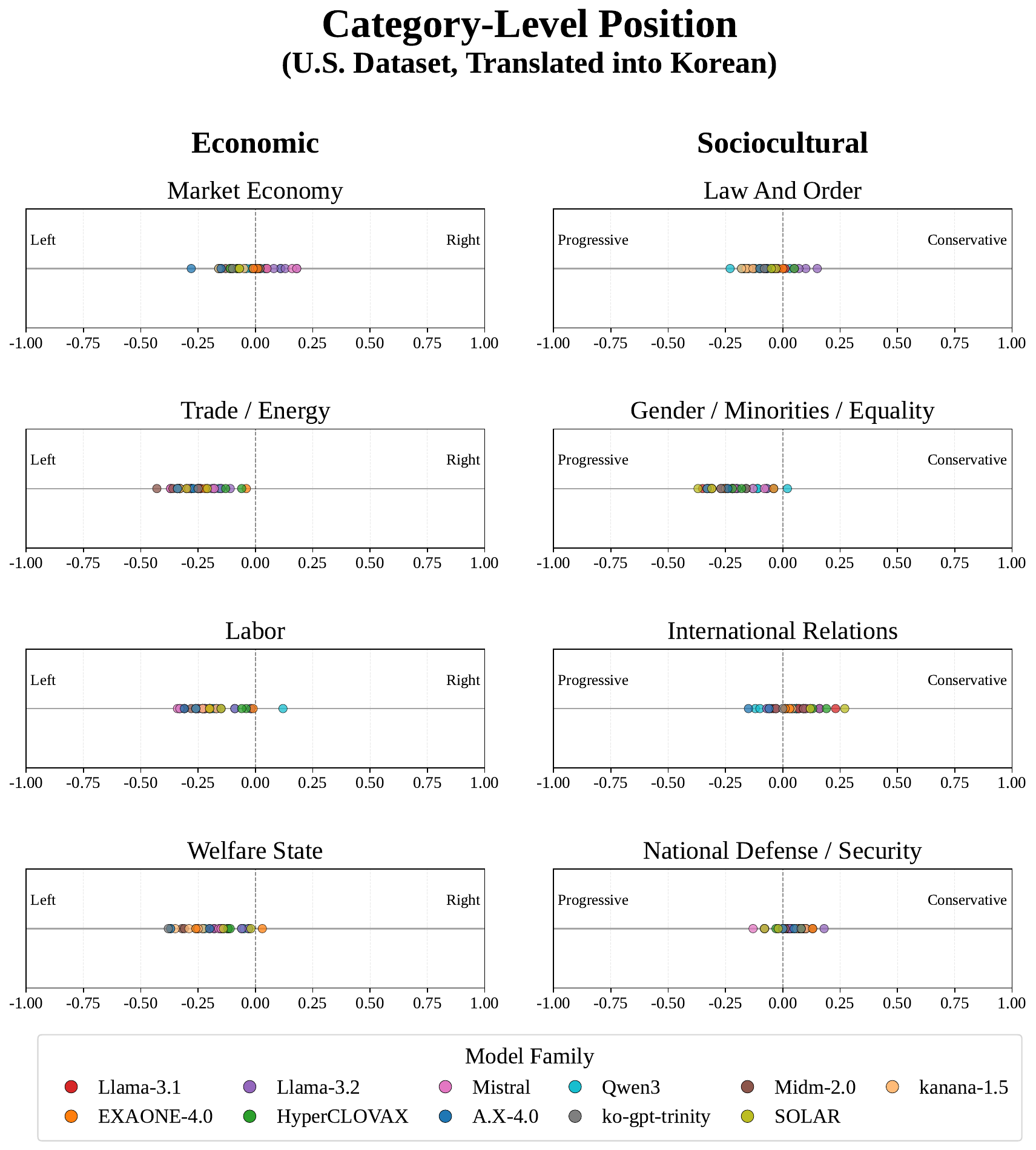}
    \caption{Category-level political positions on the U.S. dataset translated into Korean. The figure shows how model preferences vary across economic and sociocultural categories under the translated input setting.}
    \label{fig:us2ko_category}
\end{figure*}

\begin{figure*}[p]
    \centering
    \includegraphics[width=1\textwidth]{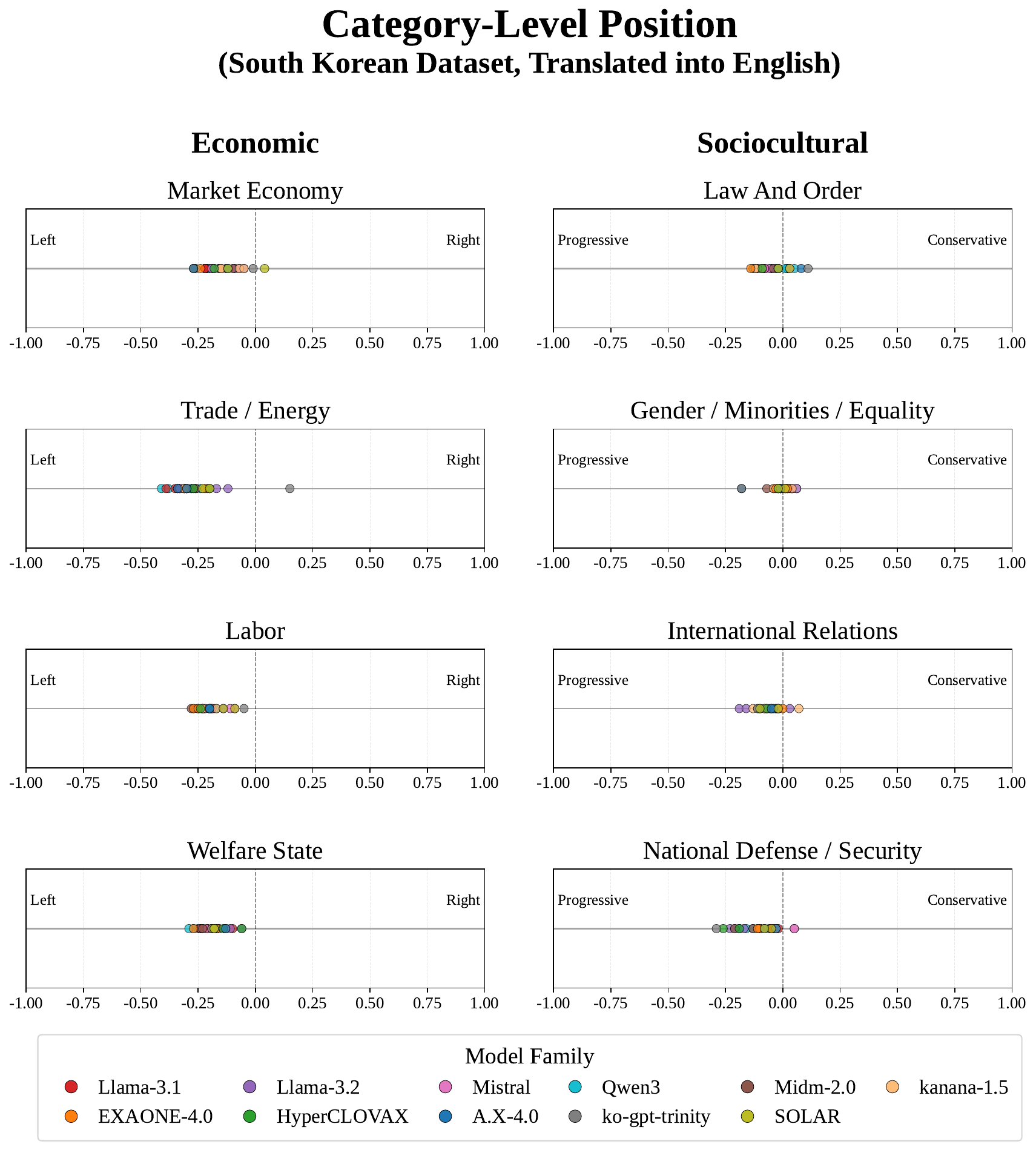}
    \caption{Category-level political positions on the South Korean dataset translated into English. The figure shows how model preferences vary across economic and sociocultural categories under the translated input setting.}
    \label{fig:ko2en_category}
\end{figure*}

Figures~\ref{fig:us_category}, \ref{fig:ko_category}, \ref{fig:us2ko_category}, and \ref{fig:ko2en_category} show category-level position distributions for the U.S., South Korean, and translated datasets. These figures provide a more detailed view of the category-level patterns discussed in the main text. In the U.S. dataset, models generally lean left or progressive across categories, but the strength of this tendency differs by issue domain. In the South Korean dataset, the direction of bias varies more across categories. For example, some categories within the same axis move in opposite directions, which can make the aggregate axis-level position appear more neutral. Table~\ref{tab:position-original-translated} provides the corresponding mean position values and standard deviations for the original and translated datasets, offering a more detailed numerical view of translation-driven shifts in model positions.

\subsection{Comparison Between Global and Korean-Specialized Models}
\label{app:model_group_position}

\begin{figure*}[p]
    \centering
    \includegraphics[width=\textwidth]{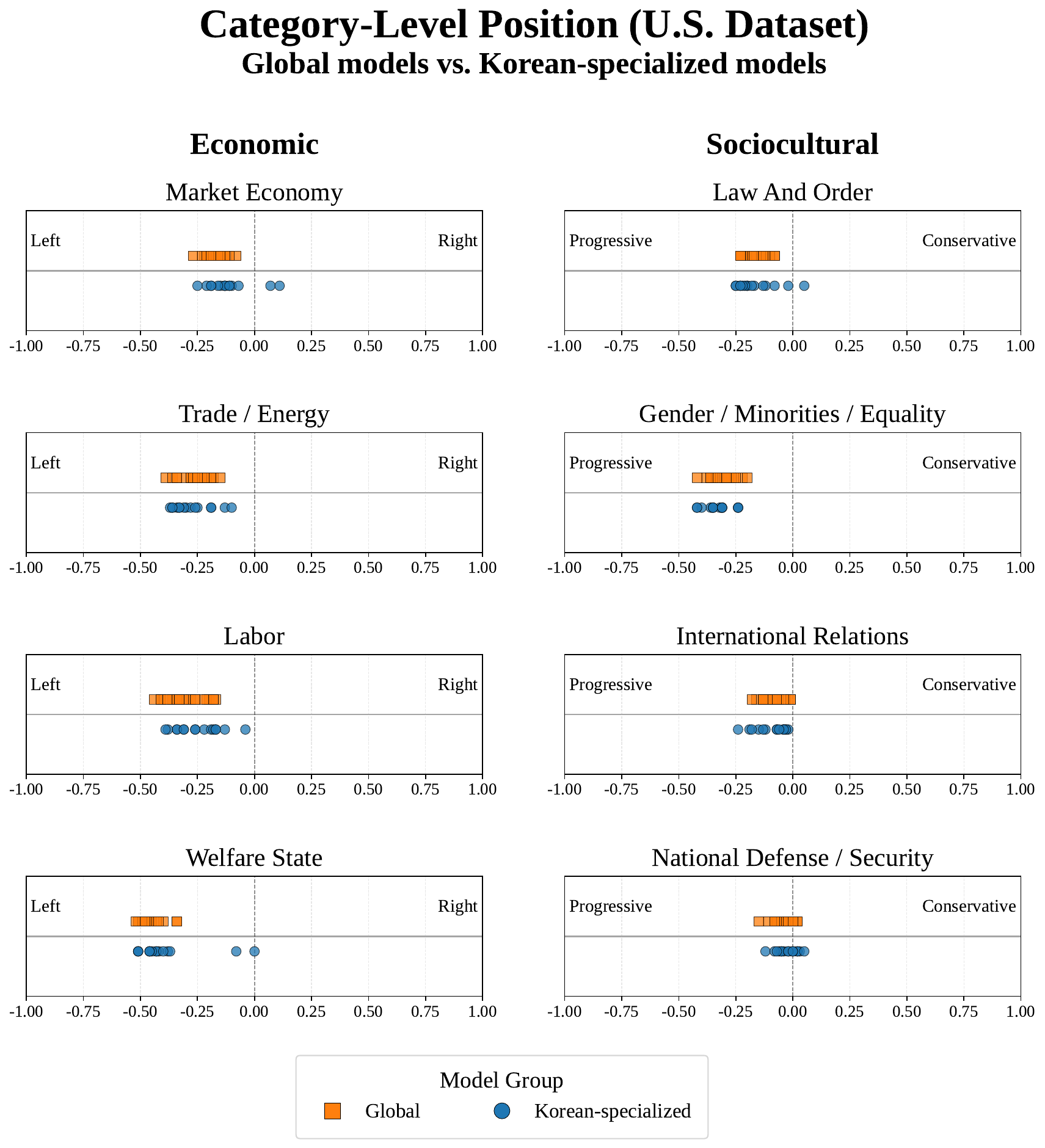}
    \caption{Category-level position distributions of global and Korean-specialized models on the U.S. dataset.}
    \label{fig:position_origin_us}
\end{figure*}

\begin{figure*}[p]
    \centering
    \includegraphics[width=\textwidth]{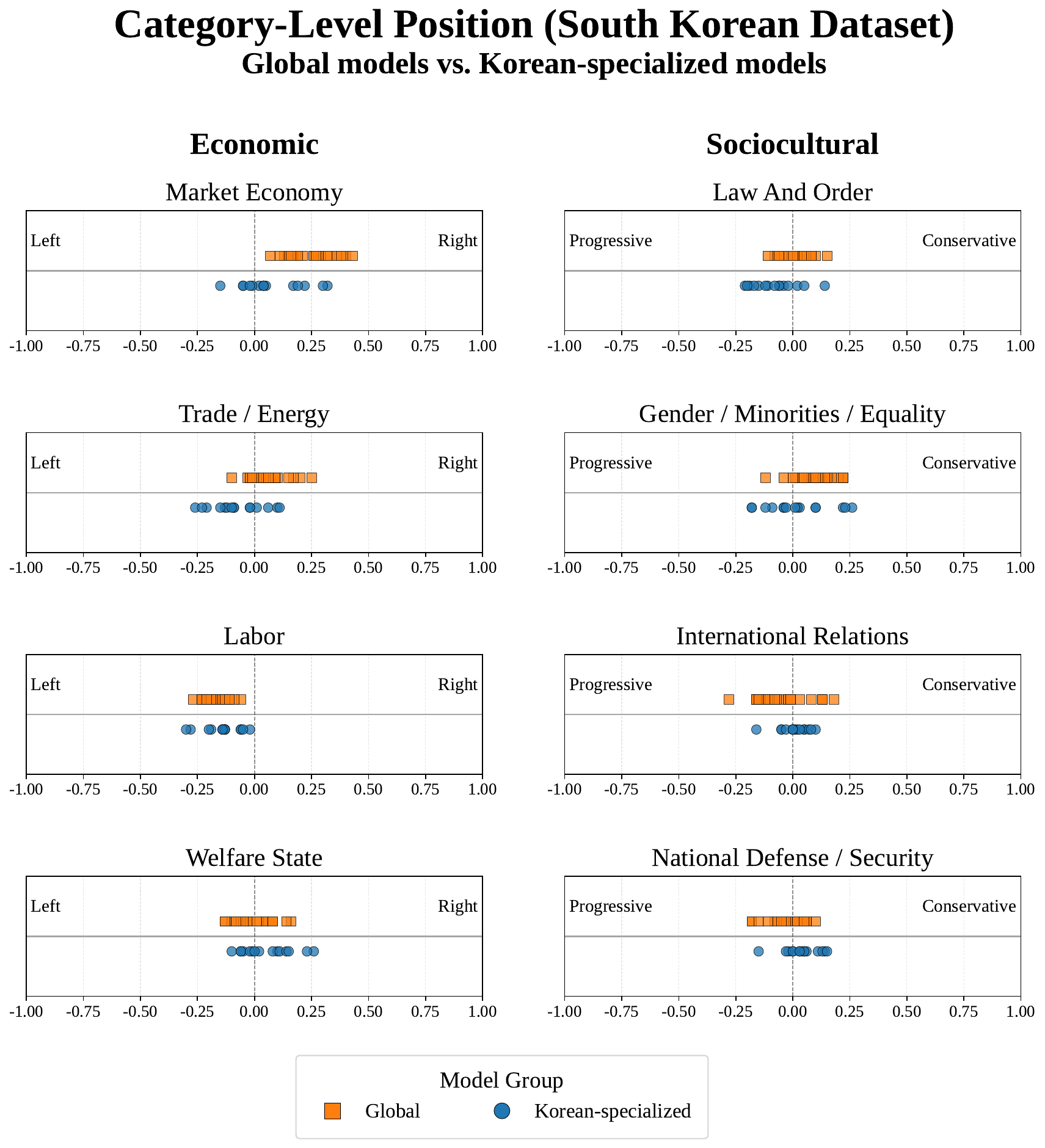}
    \caption{Category-level position distributions of global and Korean-specialized models on the South Korean dataset.}
    \label{fig:position_origin_ko}
\end{figure*}

Figures~\ref{fig:position_origin_us} and \ref{fig:position_origin_ko} compare category-level position distributions between global and Korean-specialized model groups. On the U.S. dataset, the two groups show broadly similar distributions across most categories. This suggests that Korean-specialized models exhibit position patterns similar to global models when evaluated in an English and U.S. political context. In contrast, the South Korean dataset shows clearer differences between the two groups in several categories. This pattern supports the main-text observation that a model's target language and deployment context matter more when the evaluation context is linguistically and politically local. Table~\ref{tab:position-model-group} provides the corresponding mean position values and standard deviations by model group, offering a more detailed numerical view of the differences shown in Figures~\ref{fig:position_origin_us} and~\ref{fig:position_origin_ko}.

\subsection{Model-Level Category Patterns}
\label{app:model_level_category_patterns}

\begin{figure*}[t]
    \centering

    \begin{subfigure}{0.92\textwidth}
        \centering
        \includegraphics[width=\textwidth]{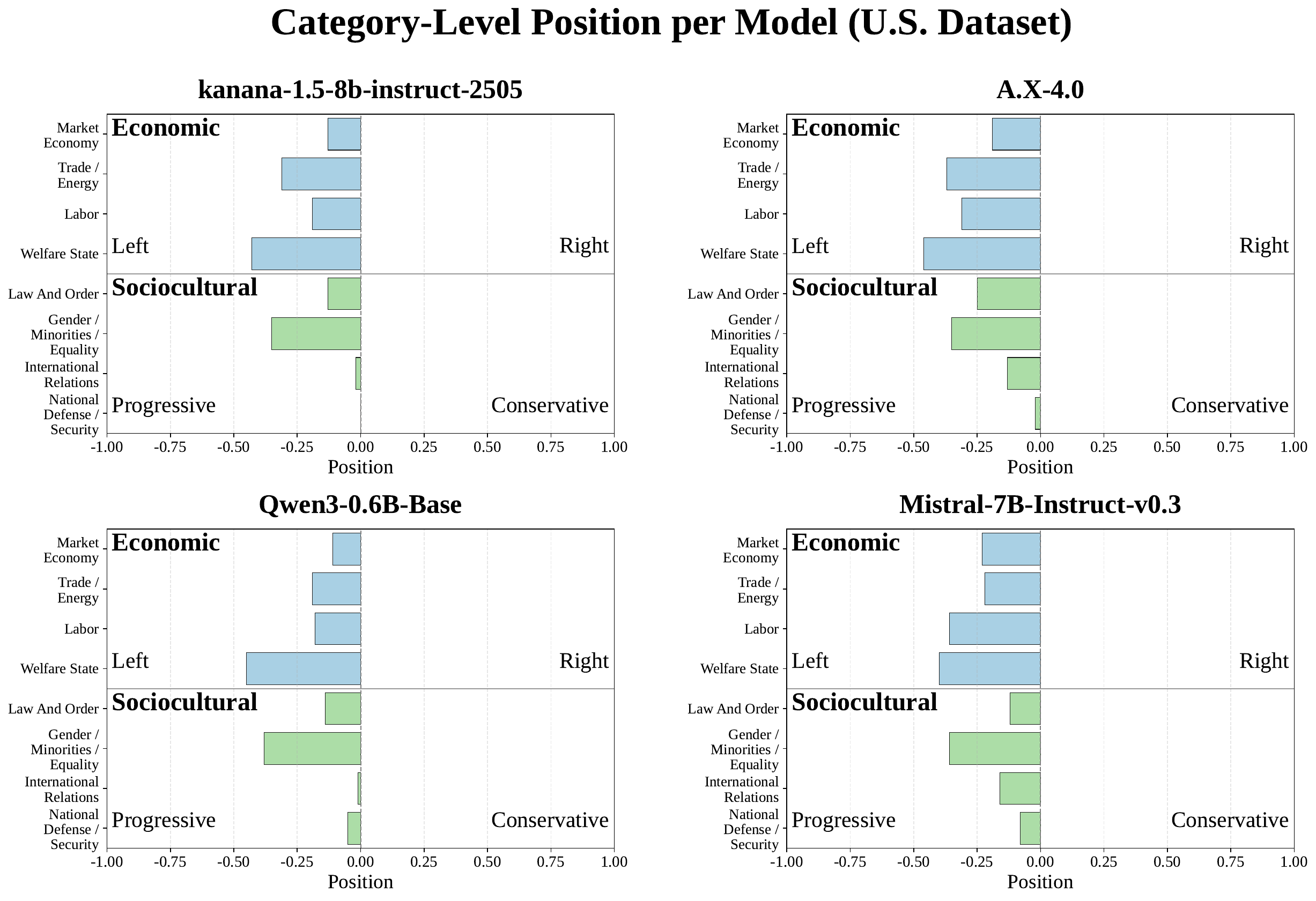}
        \caption{U.S. dataset}
        \label{fig:position_category_example_us}
    \end{subfigure}

    \vspace{0.6em}

    \begin{subfigure}{0.92\textwidth}
        \centering
        \includegraphics[width=\textwidth]{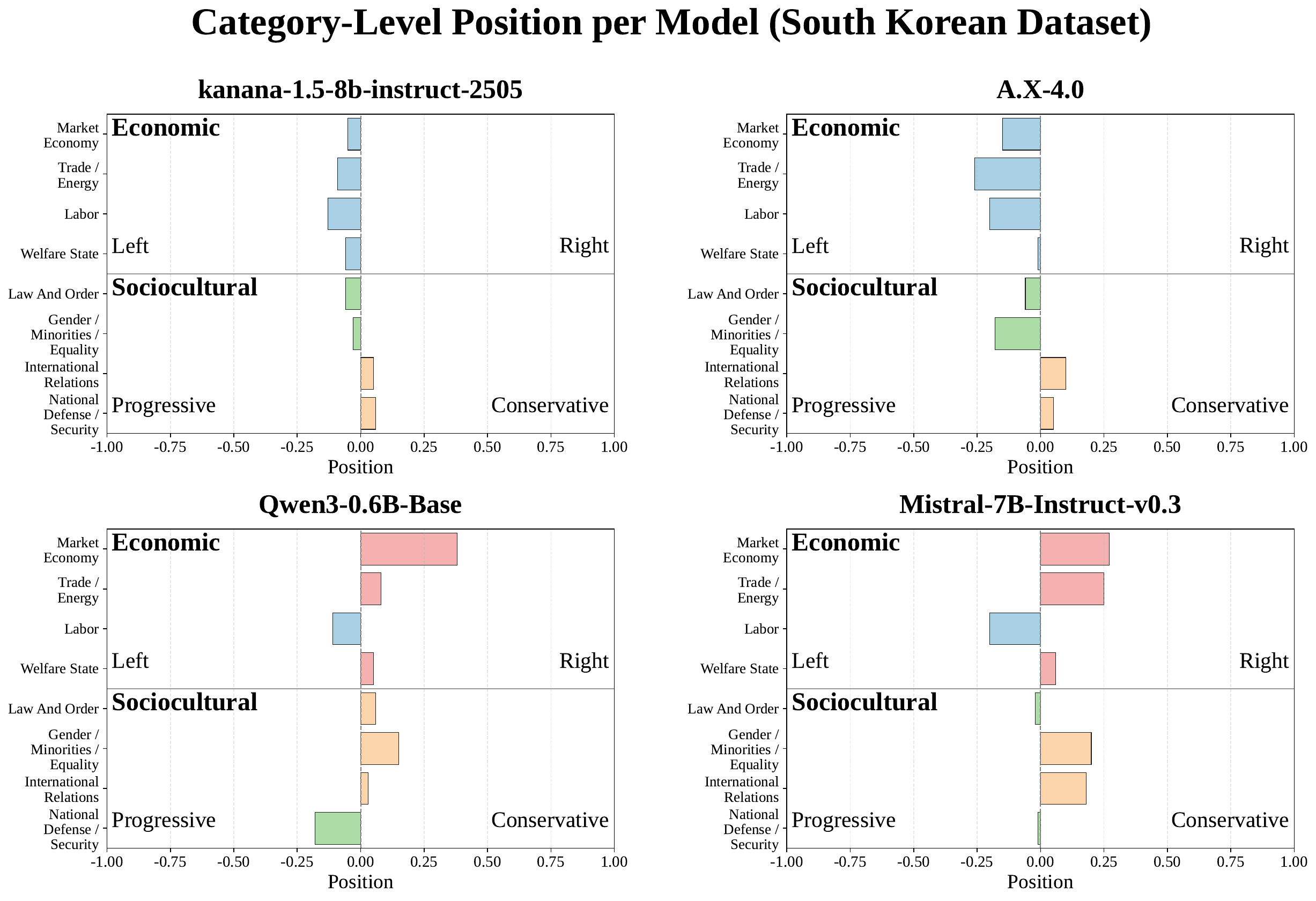}
        \caption{South Korean dataset}
        \label{fig:position_category_example_ko}
    \end{subfigure}

    \caption{Category-level political positions for representative models on the U.S. and South Korean datasets. On the U.S. dataset, models show broadly similar category-level patterns, with consistently left/progressive-leaning positions across most categories. In contrast, the South Korean dataset exhibits more heterogeneous patterns across models and categories, indicating that political bias is less uniform and more sensitive to issue context in the South Korean political setting.}
    \label{fig:model-category-examples}
\end{figure*}

Figures~\ref{fig:position_category_example_us} and \ref{fig:position_category_example_ko} provide a model-level view of the group-level patterns shown above. We select four example models and compare their category-level positions on the U.S. and South Korean datasets. On the U.S. dataset, the selected models show similar category-level patterns, with stronger bias appearing in similar categories, such as \textit{Welfare State} and \textit{Gender / Minorities / Equality}. On the South Korean dataset, the same models show more varied category-level positions. This result further indicates that model behavior becomes more heterogeneous when the evaluation context reflects local language and political discourse.

\subsection{Model-Level Metrics}
\label{app:model_level_metrics}

Tables~\ref{tab:us_total}, \ref{tab:ko_total}, \ref{tab:us2ko_total}, and \ref{tab:ko2en_total} report model-level metrics for each dataset. For each model, we report political position, neutrality score (NS), language modeling score (LMS), and ICAT score for the economic and sociocultural axes. We also report the total ICAT score, computed as the average of the two axis-level ICAT scores. These tables provide the full numerical results underlying the aggregate patterns reported in the main text, including the comparison between original and translated evaluation settings.

\subsection{Category-Level Metrics}
\label{app:category_level_metrics}

Tables~\ref{tab:us_category}, \ref{tab:ko_category}, \ref{tab:us2ko_category}, and \ref{tab:ko2en_category} report category-level metrics for each dataset. For each category, we compute position, NS, LMS, and ICAT separately. These tables show where each model exhibits stronger or weaker directional preferences and whether low ICAT scores are driven by lower neutrality or lower LMS. They also provide additional evidence that model bias can differ substantially across issue categories and language forms.

\clearpage

\input{tables/appendix/mean_position_translation}

\clearpage
\input{tables/appendix/mean_position_origin}

\clearpage

\input{tables/appendix/us_total}
\input{tables/appendix/ko_total}
\input{tables/appendix/us2ko_total}
\input{tables/appendix/ko2en_total}

\input{tables/appendix/us_category}
\input{tables/appendix/ko_category}
\input{tables/appendix/us2ko_category}
\input{tables/appendix/ko2en_category}

%% file: tables/appendix/polar_taxonomy.tex
\begin{table*}[!t]
\centering
\footnotesize
\setlength{\tabcolsep}{4pt}
\renewcommand{\arraystretch}{1.12}

\begin{tabularx}{\textwidth}{
    >{\raggedright\arraybackslash}p{0.10\textwidth}
    >{\raggedright\arraybackslash}p{0.18\textwidth}
    >{\raggedright\arraybackslash}p{0.11\textwidth}
    >{\centering\arraybackslash}p{0.13\textwidth}
    >{\raggedright\arraybackslash}X
}
\toprule
\textbf{Axis} & \textbf{Category} & \textbf{Position} & \textbf{Code} & \textbf{Subcategory} \\
\midrule

\multirow[t]{16}{=}{\textbf{Economic}}
& \multirow[t]{7}{=}{Market Economy}
& \multirow[t]{4}{=}{Left}
& 403 & Market Regulation \\
& & & 405 & Corporatism / Mixed Economy \\
& & & 412 & Controlled Economy \\
& & & 413 & Nationalization \\
\cmidrule(lr){3-5}
& & \multirow[t]{3}{=}{Right}
& 401 & Free Market Economy \\
& & & 402 & Incentives: Positive \\
& & & 414 & Economic Orthodoxy \\
\cmidrule(lr){2-5}

& \multirow[t]{3}{=}{Trade / Energy}
& \multirow[t]{2}{=}{Left}
& 406 & Protectionism: Positive \\
& & & 501 & Environmental Protection \\
\cmidrule(lr){3-5}
& & Right
& 407 & Protectionism: Negative \\
\cmidrule(lr){2-5}

& \multirow[t]{2}{=}{Labor}
& Left
& 701 & Labor Groups: Positive \\
\cmidrule(lr){3-5}
& & Right
& 702 & Labor Groups: Negative \\
\cmidrule(lr){2-5}

& \multirow[t]{4}{=}{Welfare State}
& \multirow[t]{2}{=}{Left}
& 504 & Welfare State Expansion \\
& & & 506 & Education Expansion \\
\cmidrule(lr){3-5}
& & \multirow[t]{2}{=}{Right}
& 505 & Welfare State Limitation \\
& & & 507 & Education Limitation \\
\midrule

\multirow[t]{20}{=}{\textbf{Sociocultural}}
& \multirow[t]{6}{=}{Law and Order}
& \multirow[t]{3}{=}{Progressive}
& 201 & Freedom and Human Rights \\
& & & 301 & Decentralization \\
& & & 602 & National Way of Life: Negative \\
\cmidrule(lr){3-5}
& & \multirow[t]{3}{=}{Conservative}
& 302 & Centralization \\
& & & 601 & National Way of Life: Positive \\
& & & 605 & Law and Order Reinforcement \\
\cmidrule(lr){2-5}

& \multirow[t]{7}{=}{Gender / Minorities / Equality}
& \multirow[t]{4}{=}{Progressive}
& 503 & Equality: Positive \\
& & & 604 & Traditional Morality: Negative \\
& & & 607 & Multiculturalism: Positive \\
& & & 705 & Underprivileged Minority Groups \\
\cmidrule(lr){3-5}
& & \multirow[t]{3}{=}{Conservative}
& 603 & Traditional Morality: Positive \\
& & & 608 & Multiculturalism: Negative \\
& & & 704 & Middle Class and Professional Groups \\
\cmidrule(lr){2-5}

& \multirow[t]{4}{=}{International Relations}
& \multirow[t]{2}{=}{Progressive}
& 101, 102 & Foreign Special Relationship \\
& & & 107 & Internationalism: Positive \\
\cmidrule(lr){3-5}
& & \multirow[t]{2}{=}{Conservative}
& 101, 102 & Foreign Special Relationship \\
& & & 109 & Internationalism: Negative \\
\cmidrule(lr){2-5}

& \multirow[t]{3}{=}{National Defense / Security}
& \multirow[t]{2}{=}{Progressive}
& 105 & Military: Negative \\
& & & 106 & Peace \\
\cmidrule(lr){3-5}
& & Conservative
& 104 & Military: Positive \\

\bottomrule
\end{tabularx}

\caption{Mapping between Polar categories and Manifesto Project coding scheme. Each Polar category is associated with Manifesto codes and directional labels used to assign political positions.}
\label{tab:polar_taxonomy}
\end{table*}

%% file: tables/appendix/bill_passage_rate.tex
\begin{table*}[t]
\centering
\small
\setlength{\tabcolsep}{5pt}
\renewcommand{\arraystretch}{1.12}

\begin{tabularx}{\textwidth}{
    >{\raggedright\arraybackslash}p{0.13\textwidth}
    >{\raggedright\arraybackslash}X
    >{\centering\arraybackslash}p{0.12\textwidth}
    >{\centering\arraybackslash}p{0.10\textwidth}
    >{\centering\arraybackslash}p{0.12\textwidth}
    >{\centering\arraybackslash}p{0.10\textwidth}
}
\toprule
& & \multicolumn{2}{c}{\textbf{U.S.}} & \multicolumn{2}{c}{\textbf{South Korea}} \\
\cmidrule(lr){3-4} \cmidrule(lr){5-6}
\textbf{Axis} & \textbf{Category} 
& \textbf{Passage Rate} & \textbf{Instances} 
& \textbf{Passage Rate} & \textbf{Instances} \\
\midrule
Economic & Market Economy & 11.22\% & 131 & 8.03\% & 129 \\
Economic & Trade / Energy & 8.84\% & 134 & 11.48\% & 125 \\
Economic & Labor & 15.32\% & 125 & 8.23\% & 128 \\
Economic & Welfare State & 11.57\% & 130 & 12.36\% & 124 \\
\midrule
Sociocultural & Law and Order & 19.26\% & 120 & 6.58\% & 132 \\
Sociocultural & Gender / Minorities / Equality & 27.55\% & 110 & 12.39\% & 125 \\
Sociocultural & International Relations & 8.27\% & 134 & 12.42\% & 122 \\
Sociocultural & National Defense / Security & 18.53\% & 120 & 13.49\% & 124 \\
\midrule
\multicolumn{2}{r}{\textbf{Total}} & & \textbf{1,004} & & \textbf{1,009} \\
\bottomrule
\end{tabularx}

\caption{Bill passage rates and instance allocation by category in Polar. Passage rates are computed from official legislative records from 2016 to 2025. Categories with lower passage rates receive more instances.}
\label{tab:bill_passage_allocation}
\end{table*}

%% file: tables/appendix/mean_position_translation.tex
\begin{table*}[t]
\centering
\small
\setlength{\tabcolsep}{5pt}
\renewcommand{\arraystretch}{1.12}

\begin{tabularx}{\textwidth}{
    >{\raggedright\arraybackslash}p{0.25\textwidth}
    >{\raggedright\arraybackslash}p{0.12\textwidth}
    >{\centering\arraybackslash}X
    >{\centering\arraybackslash}X
}
\toprule
& & \multicolumn{2}{c}{\textbf{Mean Position (Standard Deviation)}} \\
\cmidrule(lr){3-4}
\textbf{Category} & \textbf{Translation} & \textbf{U.S. Dataset} & \textbf{South Korean Dataset} \\
\midrule

\multirow[t]{2}{=}{\textbf{Total}\\\textit{Economic, Sociocultural}}
& Original 
& $-0.28$ ($\pm 0.07$), $-0.15$ ($\pm 0.04$) 
& $0.02$ ($\pm 0.07$), $0.01$ ($\pm 0.05$) \\
& Translated 
& $-0.16$ ($\pm 0.08$), $-0.04$ ($\pm 0.05$) 
& $-0.20$ ($\pm 0.05$), $-0.05$ ($\pm 0.03$) \\
\midrule

\multirow[t]{2}{=}{Market Economy}
& Original 
& $-0.16$ ($\pm 0.07$) 
& $0.18$ ($\pm 0.13$) \\
& Translated 
& $-0.03$ ($\pm 0.10$) 
& $-0.16$ ($\pm 0.08$) \\
\cmidrule(lr){1-4}

\multirow[t]{2}{=}{Trade / Energy}
& Original 
& $-0.26$ ($\pm 0.07$) 
& $0.01$ ($\pm 0.11$) \\
& Translated 
& $-0.25$ ($\pm 0.09$) 
& $-0.27$ ($\pm 0.09$) \\
\cmidrule(lr){1-4}

\multirow[t]{2}{=}{Labor}
& Original 
& $-0.28$ ($\pm 0.09$) 
& $-0.16$ ($\pm 0.06$) \\
& Translated 
& $-0.19$ ($\pm 0.10$) 
& $-0.20$ ($\pm 0.05$) \\
\cmidrule(lr){1-4}

\multirow[t]{2}{=}{Welfare State}
& Original 
& $-0.42$ ($\pm 0.10$) 
& $0.03$ ($\pm 0.09$) \\
& Translated 
& $-0.18$ ($\pm 0.10$) 
& $-0.18$ ($\pm 0.05$) \\
\midrule

\multirow[t]{2}{=}{Law and Order}
& Original 
& $-0.19$ ($\pm 0.07$) 
& $-0.02$ ($\pm 0.09$) \\
& Translated 
& $-0.07$ ($\pm 0.08$) 
& $-0.05$ ($\pm 0.06$) \\
\cmidrule(lr){1-4}

\multirow[t]{2}{=}{Gender / Minorities / Equality}
& Original 
& $-0.33$ ($\pm 0.06$) 
& $0.08$ ($\pm 0.11$) \\
& Translated 
& $-0.21$ ($\pm 0.10$) 
& $-0.01$ ($\pm 0.05$) \\
\cmidrule(lr){1-4}

\multirow[t]{2}{=}{International Relations}
& Original 
& $-0.08$ ($\pm 0.06$) 
& $-0.03$ ($\pm 0.10$) \\
& Translated 
& $0.04$ ($\pm 0.09$) 
& $-0.06$ ($\pm 0.05$) \\
\cmidrule(lr){1-4}

\multirow[t]{2}{=}{National Defense / Security}
& Original 
& $-0.03$ ($\pm 0.04$) 
& $0.01$ ($\pm 0.08$) \\
& Translated 
& $0.04$ ($\pm 0.06$) 
& $-0.11$ ($\pm 0.08$) \\

\bottomrule
\end{tabularx}

\caption{Mean political positions and standard deviations for the original and translated datasets. Negative values indicate left or progressive preferences, while positive values indicate right or conservative preferences.}
\label{tab:position-original-translated}
\end{table*}

%% file: tables/appendix/mean_position_origin.tex
\begin{table*}[t]
\centering
\small
\setlength{\tabcolsep}{3pt}
\renewcommand{\arraystretch}{1.12}

\begin{tabularx}{\textwidth}{
    >{\raggedright\arraybackslash}p{0.23\textwidth}
    >{\raggedright\arraybackslash}p{0.17\textwidth}
    >{\centering\arraybackslash}p{0.285\textwidth}
    >{\centering\arraybackslash}p{0.285\textwidth}
}
\toprule
& & \multicolumn{2}{c}{\textbf{Mean Position (Standard Deviation)}} \\
\cmidrule(lr){3-4}
\textbf{Category} & \textbf{Model Group} & \textbf{U.S. Dataset} & \textbf{South Korean Dataset} \\
\midrule

\multirow[t]{2}{=}{\textbf{Total}\\\textit{Economic, Sociocultural}}
& Global 
& \mbox{$-0.29$ ($\pm 0.04$), $-0.15$ ($\pm 0.02$)}
& \mbox{$0.04$ ($\pm 0.05$), $0.02$ ($\pm 0.03$)} \\
& \mbox{Korean-specialized}
& \mbox{$-0.26$ ($\pm 0.09$), $-0.15$ ($\pm 0.05$)}
& \mbox{$-0.02$ ($\pm 0.08$), $0.00$ ($\pm 0.06$)} \\
\midrule

\multirow[t]{2}{=}{Market Economy}
& Global 
& $-0.18$ ($\pm 0.04$) 
& $0.25$ ($\pm 0.09$) \\
& \mbox{Korean-specialized}
& $-0.13$ ($\pm 0.09$) 
& $0.08$ ($\pm 0.12$) \\
\cmidrule(lr){1-4}

\multirow[t]{2}{=}{Trade / Energy}
& Global 
& $-0.25$ ($\pm 0.06$) 
& $0.07$ ($\pm 0.08$) \\
& \mbox{Korean-specialized}
& $-0.27$ ($\pm 0.09$) 
& $-0.07$ ($\pm 0.11$) \\
\cmidrule(lr){1-4}

\multirow[t]{2}{=}{Labor}
& Global 
& $-0.30$ ($\pm 0.08$) 
& $-0.17$ ($\pm 0.05$) \\
& \mbox{Korean-specialized}
& $-0.25$ ($\pm 0.10$) 
& $-0.14$ ($\pm 0.08$) \\
\cmidrule(lr){1-4}

\multirow[t]{2}{=}{Welfare State}
& Global 
& $-0.44$ ($\pm 0.05$) 
& $0.01$ ($\pm 0.08$) \\
& \mbox{Korean-specialized}
& $-0.38$ ($\pm 0.14$) 
& $0.06$ ($\pm 0.10$) \\
\midrule

\multirow[t]{2}{=}{Law and Order}
& Global 
& $-0.19$ ($\pm 0.05$) 
& $0.02$ ($\pm 0.06$) \\
& \mbox{Korean-specialized}
& $-0.18$ ($\pm 0.09$) 
& $-0.08$ ($\pm 0.09$) \\
\cmidrule(lr){1-4}

\multirow[t]{2}{=}{Gender / Minorities / Equality}
& Global 
& $-0.33$ ($\pm 0.05$) 
& $0.11$ ($\pm 0.08$) \\
& \mbox{Korean-specialized}
& $-0.34$ ($\pm 0.07$) 
& $0.04$ ($\pm 0.14$) \\
\cmidrule(lr){1-4}

\multirow[t]{2}{=}{International Relations}
& Global 
& $-0.08$ ($\pm 0.05$) 
& $-0.04$ ($\pm 0.11$) \\
& \mbox{Korean-specialized}
& $-0.08$ ($\pm 0.07$) 
& $0.00$ ($\pm 0.06$) \\
\cmidrule(lr){1-4}

\multirow[t]{2}{=}{National Defense / Security}
& Global 
& $-0.03$ ($\pm 0.04$) 
& $0.00$ ($\pm 0.07$) \\
& \mbox{Korean-specialized}
& $-0.01$ ($\pm 0.05$) 
& $0.03$ ($\pm 0.08$) \\

\bottomrule
\end{tabularx}

\caption{Mean political positions and standard deviations by model group. Global and Korean-specialized models are compared across the U.S. and Korean datasets. Negative values indicate left or progressive preferences, while positive values indicate right or conservative preferences.}
\label{tab:position-model-group}
\end{table*}

%% file: tables/appendix/us_total.tex
\begin{table*}[htbp]
\centering
\resizebox{\textwidth}{!}{%
\begin{tabular}{@{}lrrrrrrrrr@{}}
\toprule
\multicolumn{1}{c}{\multirow{2}{*}{\textbf{Model}}} &
  \multicolumn{4}{c}{\textbf{Economic}} &
  \multicolumn{4}{c}{\textbf{Sociocultural}} &
  \multicolumn{1}{c}{\textbf{Total}} \\ \cmidrule(l){2-5}\cmidrule(l){6-9}
\multicolumn{1}{c}{} &
  \multicolumn{1}{c}{\textbf{Position}} &
  \multicolumn{1}{c}{\textbf{NS}} &
  \multicolumn{1}{c}{\textbf{LMS}} &
  \multicolumn{1}{c}{\textbf{ICAT}} &
  \multicolumn{1}{c}{\textbf{Position}} &
  \multicolumn{1}{c}{\textbf{NS}} &
  \multicolumn{1}{c}{\textbf{LMS}} &
  \multicolumn{1}{c}{\textbf{ICAT}} &
  \multicolumn{1}{c}{\textbf{ICAT}} \\ \midrule \midrule
Llama-3.1-8B                        & -0.34 & 0.66 & 98.46          & 64.53 & -0.15 & 0.84 & 98.55          & 83.18          & 73.86 \\
Llama-3.1-8B-Instruct               & -0.30 & 0.70 & 98.08          & 68.35 & -0.20 & 0.79 & 97.93          & 77.27          & 72.81 \\
Llama-3.1-70B                       & -0.34 & 0.66 & 97.88          & 64.28 & -0.14 & 0.85 & 98.76          & 84.22          & 74.25 \\
Llama-3.1-70B-Instruct              & -0.33 & 0.67 & 98.46          & 66.31 & -0.18 & 0.81 & 98.55          & 80.20          & 73.25 \\
Llama-3.2-1B                        & -0.28 & 0.72 & \textbf{98.85} & 71.23 & -0.11 & 0.88 & 98.76          & \textbf{86.59} & 78.91 \\
Llama-3.2-1B-Instruct               & -0.27 & 0.73 & 96.92          & 70.47 & -0.13 & 0.85 & 96.28          & 81.92          & 76.20 \\
\bottomrule
\end{tabular}%
}
\caption{Model-level evaluation results on the U.S. dataset.}
\label{tab:us_total}
\end{table*}

%% file: tables/appendix/ko_total.tex
\begin{table*}[htbp]
\centering
\resizebox{\textwidth}{!}{%
\begin{tabular}{@{}lrrrrrrrrr@{}}
\toprule    
\multicolumn{1}{c}{\multirow{2}{*}{\textbf{Model}}} &
  \multicolumn{4}{c}{\textbf{Economic}} & 
  \multicolumn{4}{c}{\textbf{Sociocultural}} &
  \multicolumn{1}{c}{\textbf{Total}} \\ \cmidrule(l){2-5}\cmidrule(l){6-9}
\multicolumn{1}{c}{} &
  \multicolumn{1}{c}{\textbf{Position}} &
  \multicolumn{1}{c}{\textbf{NS}} &
  \multicolumn{1}{c}{\textbf{LMS}} &
  \multicolumn{1}{c}{\textbf{ICAT}} &
  \multicolumn{1}{c}{\textbf{Position}} &
  \multicolumn{1}{c}{\textbf{NS}} &
  \multicolumn{1}{c}{\textbf{LMS}} &
  \multicolumn{1}{c}{\textbf{ICAT}} &
  \multicolumn{1}{c}{\textbf{ICAT}} \\ \midrule\midrule
Llama-3.1-8B                        & 0.00  & 0.83 & 96.84 & 80.55 & 0.01 & 0.91 & 97.22  & 88.24 & 84.39 \\
Llama-3.1-8B-Instruct               & 0.06  & 0.84 & 94.47 & 78.97 & 0.01 & 0.91 & 95.63  & 86.91 & 82.94 \\
Llama-3.1-70B                       & -0.03 & 0.88 & 98.02 & 86.10 & -0.02 & 0.97 & 98.61  & \textbf{96.03} & 91.07 \\
Llama-3.1-70B-Instruct              & -0.03 & 0.86 & 97.83 & 84.40 & -0.01 & 0.94 & 97.81  & 91.87 & 88.13 \\
Llama-3.2-1B                        & 0.07  & 0.84 & 90.91 & 76.19 & 0.01  & 0.95 & 93.04  & 88.24 & 82.22 \\
Llama-3.2-1B-Instruct               & 0.09  & 0.84 & 86.56 & 72.66 & -0.02 & 0.85 & 86.08  & 72.91 & 72.78 \\
\bottomrule
\end{tabular}%
}
\caption{Model-level evaluation results on the South Korean dataset.}
\label{tab:ko_total}
\end{table*}

%% file: tables/appendix/us2ko_total.tex
\begin{table*}[htbp]
\centering
\resizebox{\textwidth}{!}{%
\begin{tabular}{@{}lrrrrrrrrr@{}}
\toprule
\multicolumn{1}{c}{\multirow{2}{*}{\textbf{Model}}} &
  \multicolumn{4}{c}{\textbf{Economic}} &
  \multicolumn{4}{c}{\textbf{Sociocultural}} &
  \multicolumn{1}{c}{\textbf{Total}} \\ \cmidrule(l){2-5}\cmidrule(l){6-9}
\multicolumn{1}{c}{} &
  \multicolumn{1}{c}{\textbf{Position}} &
  \multicolumn{1}{c}{\textbf{NS}} &
  \multicolumn{1}{c}{\textbf{LMS}} &
  \multicolumn{1}{c}{\textbf{ICAT}} &
  \multicolumn{1}{c}{\textbf{Position}} &
  \multicolumn{1}{c}{\textbf{NS}} &
  \multicolumn{1}{c}{\textbf{LMS}} &
  \multicolumn{1}{c}{\textbf{ICAT}} &
  \multicolumn{1}{c}{\textbf{ICAT}} \\ \midrule\midrule
Llama-3.1-8B                        & -0.16 & 0.84          & 95.96          & 80.97          & 0.02  & 0.84          & 94.21          & 79.59          & 80.28          \\
Llama-3.1-8B-Instruct               & -0.14 & 0.83          & 93.46          & 77.42          & 0.00  & 0.90          & 92.98          & 83.51          & 80.47          \\
Llama-3.1-70B                       & -0.23 & 0.76          & 95.77          & 73.26          & -0.06 & 0.87          & 94.63          & 82.79          & 78.03          \\
Llama-3.1-70B-Instruct              & -0.21 & 0.79          & 95.00          & 75.44          & -0.09 & 0.87          & 93.60          & 81.34          & 78.39          \\
Llama-3.2-1B                        & -0.05 & 0.90          & 91.54          & 82.28          & 0.06  & 0.91          & 89.46          & 81.40          & 81.84          \\
Llama-3.2-1B-Instruct               & -0.01 & 0.91          & 89.23          & 80.91          & 0.02  & 0.87          & 85.95          & 74.85          & 77.88          \\
\bottomrule
\end{tabular}%
}
\caption{Model-level evaluation results on the U.S. dataset translated into Korean.}
\label{tab:us2ko_total}
\end{table*}

%% file: tables/appendix/ko2en_total.tex
\begin{table*}[htbp]
\centering
\resizebox{\textwidth}{!}{%
\begin{tabular}{@{}lrrrrrrrrr@{}}
\toprule
\multicolumn{1}{c}{\multirow{2}{*}{\textbf{Model}}} &
  \multicolumn{4}{c}{\textbf{Economic}} &
  \multicolumn{4}{c}{\textbf{Sociocultural}} &
  \multicolumn{1}{c}{\textbf{Total}} \\ \cmidrule(l){2-5}\cmidrule(l){6-9}
\multicolumn{1}{c}{} &
  \multicolumn{1}{c}{\textbf{Position}} &
  \multicolumn{1}{c}{\textbf{NS}} &
  \multicolumn{1}{c}{\textbf{LMS}} &
  \multicolumn{1}{c}{\textbf{ICAT}} &
  \multicolumn{1}{c}{\textbf{Position}} &
  \multicolumn{1}{c}{\textbf{NS}} &
  \multicolumn{1}{c}{\textbf{LMS}} &
  \multicolumn{1}{c}{\textbf{ICAT}} &
  \multicolumn{1}{c}{\textbf{ICAT}} \\ \midrule\midrule
Llama-3.1-8B                        & -0.24 & 0.76          & 95.85          & 72.51          & -0.06 & 0.94          & 96.22          & 90.36          & 81.44          \\
Llama-3.1-8B-Instruct               & -0.21 & 0.79          & 95.26          & 75.71          & -0.05 & 0.94          & 95.43          & 89.99          & 82.85          \\
Llama-3.1-70B                       & -0.28 & 0.72          & 96.44          & 69.33          & -0.05 & 0.95          & 97.81          & 92.59          & 80.96          \\
Llama-3.1-70B-Instruct              & -0.25 & 0.75          & 95.65          & 71.43          & -0.06 & 0.93          & 97.61          & 90.49          & 80.96          \\
Llama-3.2-1B                        & -0.16 & 0.84          & 94.86          & 79.86          & -0.10 & 0.88          & 94.63          & 83.43          & 81.64          \\
Llama-3.2-1B-Instruct               & -0.11 & 0.89          & 89.72          & 79.77          & -0.11 & 0.87          & 91.65          & 79.37          & 79.57          \\
\bottomrule
\end{tabular}%
}
\caption{Model-level evaluation results on the South Korean dataset translated into English.}
\label{tab:ko2en_total}
\end{table*}

%% file: tables/appendix/us_category.tex
\onecolumn
{\footnotesize
\setlength{\tabcolsep}{3pt}
\begin{longtable}[c]{@{}lllrrrr@{}}
\caption{Category-level evaluation results on the U.S. dataset.}
\label{tab:us_category}\\*
\toprule
\multicolumn{1}{c}{\textbf{Model}} & \multicolumn{1}{c}{\textbf{Axis}} & \multicolumn{1}{c}{\textbf{Category}} & \multicolumn{1}{c}{\textbf{Position}} & \multicolumn{1}{c}{\textbf{NS}} & \multicolumn{1}{c}{\textbf{LMS}} & \multicolumn{1}{c}{\textbf{ICAT}} \\* \midrule
\endfirsthead
\multicolumn{7}{c}%
{{\bfseries Table \thetable\ continued from previous page}} \\*
\toprule
\multicolumn{1}{c}{\textbf{Model}} & \multicolumn{1}{c}{\textbf{Axis}} & \multicolumn{1}{c}{\textbf{Category}} & \multicolumn{1}{c}{\textbf{Position}} & \multicolumn{1}{c}{\textbf{NS}} & \multicolumn{1}{c}{\textbf{LMS}} & \multicolumn{1}{c}{\textbf{ICAT}} \\* \midrule
\endhead
\bottomrule
\endfoot
\endlastfoot
\multirow{8}{*}{Llama-3.1-8B} & \multirow{4}{*}{Economic} & Market Economy & -0.21 & 0.79 & 96.95 & 76.23 \\*
 &  & Trade / Energy & -0.34 & 0.66 & 99.25 & 65.18 \\*
 &  & Labor & -0.34 & 0.66 & 99.20 & 65.08 \\*
 &  & Welfare State & -0.49 & 0.51 & 100.00 & 50.77 \\* \cmidrule(l){2-7} 
 & \multirow{4}{*}{Sociocultural} & Law and Order & -0.22 & 0.78 & 98.33 & 77.03 \\*
 &  & Gender / Minorities / Equality & -0.26 & 0.74 & 96.36 & 70.96 \\*
 &  & International Relations & -0.08 & 0.92 & 100.00 & 91.79 \\*
 &  & National Defense / Security & -0.08 & 0.93 & 98.33 & 90.96 \\
\bottomrule
\end{longtable}
}
\twocolumn

%% file: tables/appendix/ko_category.tex
\onecolumn
{\footnotesize
\setlength{\tabcolsep}{3pt}
\begin{longtable}[c]{@{}lllrrrr@{}}
\caption{Category-level evaluation results on the South Korean dataset.}
\label{tab:ko_category}\\*
\toprule
\multicolumn{1}{c}{\textbf{Model}} &
  \multicolumn{1}{c}{\textbf{Axis}} &
  \multicolumn{1}{c}{\textbf{Category}} &
  \multicolumn{1}{c}{\textbf{Position}} &
  \multicolumn{1}{c}{\textbf{NS}} &
  \multicolumn{1}{c}{\textbf{LMS}} &
  \multicolumn{1}{c}{\textbf{ICAT}} \\* \midrule
\endfirsthead
\multicolumn{7}{c}%
{{\bfseries Table \thetable\ continued from previous page}} \\*
\toprule
\multicolumn{1}{c}{\textbf{Model}} &
  \multicolumn{1}{c}{\textbf{Axis}} &
  \multicolumn{1}{c}{\textbf{Category}} &
  \multicolumn{1}{c}{\textbf{Position}} &
  \multicolumn{1}{c}{\textbf{NS}} &
  \multicolumn{1}{c}{\textbf{LMS}} &
  \multicolumn{1}{c}{\textbf{ICAT}} \\* \midrule
\endhead
\bottomrule
\endfoot
\endlastfoot
\multirow{8}{*}{Llama-3.1-8B}                        & \multirow{4}{*}{Economic}      & Market Economy                 & 0.29  & 0.71 & 99.22  & 70.76  \\*
                                                     &                                & Trade / Energy                 & 0.02  & 0.98 & 94.40  & 92.13  \\*
                                                     &                                & Labor                          & -0.23 & 0.77 & 97.66  & 74.77  \\*
                                                     &                                & Welfare State                  & -0.10 & 0.90 & 96.77  & 87.41  \\* \cmidrule(l){2-7} 
                                                     & \multirow{4}{*}{Sociocultural} & Law and Order                  & -0.05 & 0.95 & 97.73  & 93.29  \\*
                                                     &                                & Gender / Minorities / Equality & 0.06  & 0.94 & 96.00  & 90.62  \\*
                                                     &                                & International Relations        & -0.12 & 0.88 & 96.72  & 84.83  \\*
                                                     &                                & National Defense / Security    & 0.06  & 0.94 & 98.39  & 92.04  \\
\bottomrule

\end{longtable}
}
\twocolumn

%% file: tables/appendix/us2ko_category.tex
\onecolumn
{\footnotesize
\setlength{\tabcolsep}{3pt}
\begin{longtable}[c]{@{}lllrrrr@{}}
\caption{Category-level evaluation results on the U.S. dataset translated into Korean.}
\label{tab:us2ko_category}\\*
\toprule
\multicolumn{1}{c}{\textbf{Model}} & \multicolumn{1}{c}{\textbf{Axis}} & \multicolumn{1}{c}{\textbf{Category}} & \multicolumn{1}{c}{\textbf{Position}} & \multicolumn{1}{c}{\textbf{NS}} & \multicolumn{1}{c}{\textbf{LMS}} & \multicolumn{1}{c}{\textbf{ICAT}} \\* \midrule
\endfirsthead
\multicolumn{7}{c}%
{{\bfseries Table \thetable\ continued from previous page}} \\*
\toprule
\multicolumn{1}{c}{\textbf{Model}} & \multicolumn{1}{c}{\textbf{Axis}} & \multicolumn{1}{c}{\textbf{Category}} & \multicolumn{1}{c}{\textbf{Position}} & \multicolumn{1}{c}{\textbf{NS}} & \multicolumn{1}{c}{\textbf{LMS}} & \multicolumn{1}{c}{\textbf{ICAT}} \\* \midrule
\endhead
\bottomrule
\endfoot
\endlastfoot
\multirow{8}{*}{Llama-3.1-8B} & \multirow{4}{*}{Economic} & Market Economy & 0.00 & 1.00 & 95.42 & 95.42 \\*
 &  & Trade / Energy & -0.21 & 0.79 & 96.27 & 76.15 \\*
 &  & Labor & -0.22 & 0.78 & 99.20 & 77.77 \\*
 &  & Welfare State & -0.20 & 0.80 & 93.08 & 74.46 \\* \cmidrule(l){2-7} 
 & \multirow{4}{*}{Sociocultural} & Law and Order & -0.02 & 0.98 & 90.00 & 88.50 \\*
 &  & Gender / Minorities / Equality & -0.27 & 0.73 & 90.00 & 65.45 \\*
 &  & International Relations & 0.23 & 0.77 & 99.25 & 76.29 \\*
 &  & National Defense / Security & 0.10 & 0.90 & 96.67 & 87.00 \\
\bottomrule
\end{longtable}
}
\twocolumn

%% file: tables/appendix/ko2en_category.tex
\onecolumn
{\footnotesize
\setlength{\tabcolsep}{3pt}
\begin{longtable}[c]{@{}lllrrrr@{}}
\caption{Category-level evaluation results on the South Korean dataset translated into English.}
\label{tab:ko2en_category}\\*
\toprule
\multicolumn{1}{c}{\textbf{Model}} & \multicolumn{1}{c}{\textbf{Axis}} & \multicolumn{1}{c}{\textbf{Category}} & \multicolumn{1}{c}{\textbf{Position}} & \multicolumn{1}{c}{\textbf{NS}} & \multicolumn{1}{c}{\textbf{LMS}} & \multicolumn{1}{c}{\textbf{ICAT}} \\* \midrule
\endfirsthead
\multicolumn{7}{c}%
{{\bfseries Table \thetable\ continued from previous page}} \\*
\toprule
\multicolumn{1}{c}{\textbf{Model}} & \multicolumn{1}{c}{\textbf{Axis}} & \multicolumn{1}{c}{\textbf{Category}} & \multicolumn{1}{c}{\textbf{Position}} & \multicolumn{1}{c}{\textbf{NS}} & \multicolumn{1}{c}{\textbf{LMS}} & \multicolumn{1}{c}{\textbf{ICAT}} \\* \midrule
\endhead
\bottomrule
\endfoot
\endlastfoot
\multirow{8}{*}{Llama-3.1-8B} & \multirow{4}{*}{Economic} & Market Economy & -0.22 & 0.78 & 96.90 & 75.12 \\*
 &  & Trade / Energy & -0.35 & 0.65 & 92.80 & 60.13 \\*
 &  & Labor & -0.19 & 0.81 & 97.66 & 79.35 \\*
 &  & Welfare State & -0.21 & 0.79 & 95.97 & 75.85 \\* \cmidrule(l){2-7} 
 & \multirow{4}{*}{Sociocultural} & Law and Order & -0.11 & 0.89 & 94.70 & 84.65 \\*
 &  & Gender / Minorities / Equality & -0.04 & 0.96 & 95.20 & 91.39 \\*
 &  & International Relations & -0.05 & 0.95 & 99.18 & 94.30 \\*
 &  & National Defense / Security & -0.05 & 0.95 & 95.97 & 91.32 \\
\bottomrule
\end{longtable}
}
\twocolumn